\documentclass{article}
\usepackage[utf8]{inputenc}
\usepackage[a4paper, left=1in, right=1in, top=1in, bottom=1in]{geometry}
\usepackage{authblk}

\usepackage{natbib}
\usepackage{mathpazo}
\usepackage{microtype}
\usepackage{graphicx}
\usepackage{subfigure}
\usepackage{booktabs}
\usepackage{hyperref}
\usepackage{amsmath}
\usepackage{amssymb}
\usepackage{mathtools}
\usepackage{amsthm}
\usepackage{mathrsfs}
\usepackage{thmtools}
\usepackage{url}
\usepackage{xspace}
\usepackage[parfill]{parskip}
\usepackage{lipsum}
\usepackage{fancyhdr}
\usepackage[many]{tcolorbox} 
\usepackage{xcolor}    

\fancypagestyle{firststyle}
{
   \fancyhf{}
   \fancyfoot[R]{\thepage\quad}
}

\theoremstyle{plain}

\theoremstyle{definition}

\theoremstyle{remark}

\makeatletter
\DeclareRobustCommand\onedot{\futurelet\@let@token\@onedot}
\def\@onedot{\ifx\@let@token.\else.\null\fi\xspace}

\makeatother

\definecolor{grass}{RGB}{127, 186, 0}
\definecolor{grey}{HTML}{737373}

\hypersetup{
    colorlinks,
    linkcolor={grass!50!black},
    citecolor={blue!50!black},
    urlcolor={grass!70!black}
}

\newtcolorbox{titlebox}[2][]{
    arc=3mm,
    lower separated=false,
    colback=white,
    colframe=grey,
    coltitle=grass!85!black,
    fonttitle=\Large\bfseries,
    boxed title style={
        size=small,
        colback=white,
        colframe=white,
    },
    enhanced,
    attach boxed title to top right={xshift=3mm, yshift=1mm-\tcboxedtitleheight},
    adjusted title=#2
}

\tcbset{textmarker/.style={%
        enhanced,
        parbox=false,boxrule=0mm,boxsep=0mm,arc=0mm,
        outer arc=0mm,left=6mm,right=3mm,top=7pt,bottom=7pt,
        toptitle=1mm,bottomtitle=1mm,oversize}}

\newtcolorbox{hintBox}{textmarker,
    borderline west={6pt}{0pt}{yellow},
    colback=yellow!10!white}
\newtcolorbox{importantBox}{textmarker,
    borderline west={6pt}{0pt}{red},
    colback=red!10!white}
\newtcolorbox{noteBox}{
    colback=green!10!white}
\newcommand{\note}[1]{\begin{noteBox} #1 \end{noteBox}}

\newcommand{\fancyfootnotetext}[2]{%
  \fancypagestyle{dingens}{%
    \fancyfoot[LO,RE]{\parbox{12cm}{\footnotesize #1}}%
  }%
  \thispagestyle{dingens}%
}

\makeatletter
\renewcommand\AB@affilsepx{, \protect\Affilfont}

\makeatother

\makeatletter
\renewcommand{\maketitle}{
    \begin{flushleft} 
    \vskip 1em 
    {\bfseries \@title \par}
    \vskip 0.5em 
    {\bfseries \small \@author \par}
    \vskip 2em 
    \end{flushleft}
}
\makeatother

\title{{\huge IGOR: \textcolor{grass!85!black}{I}mage-\textcolor{grass!85!black}{GO}al \textcolor{grass!85!black}{R}epresentations} \vskip 0.2em \par
{\Large Atomic Control Units for Foundation Models in Embodied AI}
}

\author[*,	$\dagger$, $\Diamond$]{Xiaoyu Chen}
\author[*, 	$\dagger$]{Junliang Guo}
\author[*, $\dagger$]{Tianyu He}
\author[*, $\dagger$]{Chuheng Zhang}
\author[$\dagger$]{Pushi Zhang}
\author[ ]{Derek Yang }
\author[*,$\dagger$]{Li Zhao}
\author[$\dagger$]{Jiang Bian}

\affil[*]{Equal contributions}
\affil[$\dagger$]{Microsoft Research}
\affil[$\Diamond$]{Tsinghua University}

\date{}

\begin{document}

\thispagestyle{firststyle}

\begin{titlebox}{\smash{\raisebox{1pt}{\includegraphics[width=1cm]{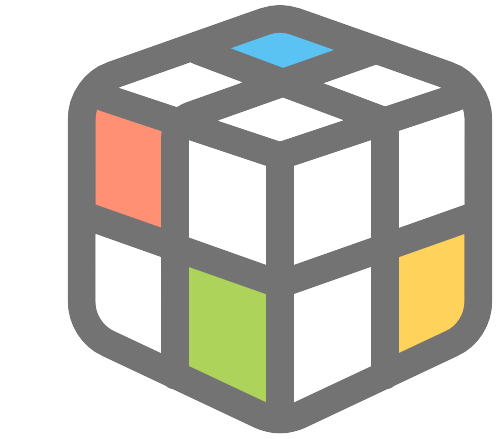}}}}
    \maketitle
    We introduce Image-GOal Representations~(IGOR), aiming to learn a unified, semantically consistent action space across human and various robots.
Through this unified latent action space, IGOR enables knowledge transfer among large-scale robot and human activity data. We achieve this by compressing visual changes between an initial image and its goal state into latent actions.
IGOR allows us to generate latent action labels for internet-scale video data. This unified latent action space enables the training of foundation policy and world models across a wide variety of tasks performed by both robots and humans. 
We demonstrate that: (1) IGOR learns a semantically consistent action space for both human and robots, characterizing various possible motions of objects representing the physical interaction knowledge;
(2) IGOR can ``migrate'' the movements of the object in the one video to other videos, even across human and robots, by jointly using the latent action model and world model;
(3) IGOR can learn to align latent actions with natural language through the foundation policy model, and integrate latent actions with a low-level policy model to achieve effective robot control.
We believe IGOR opens new possibilities for human-to-robot knowledge transfer and control.

    \vskip 0.5em 
    \hfill\begin{minipage}{0.5\linewidth}
        \ttfamily 
        Website: \url{https://aka.ms/project-igor}
    \end{minipage}
\end{titlebox}

\begin{figure*}[htb!]
    \centering
    \includegraphics[width=1.0\textwidth]{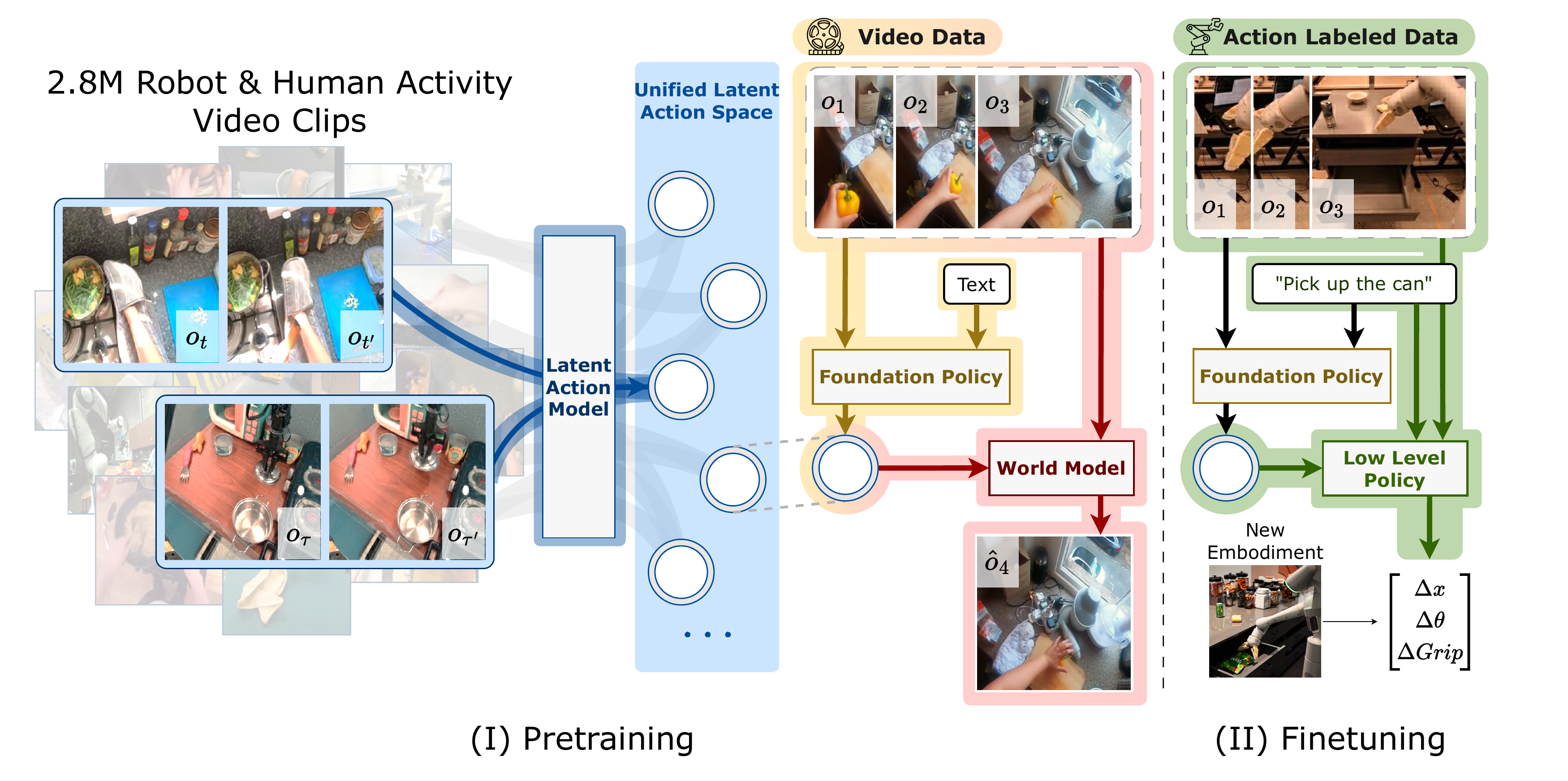}
    \caption{
    \textbf{Image-GOal Representations~(IGOR) based training framework for embodied AI. } 
    IGOR learns a unified latent action space for humans and robots by compressing visual changes between an image and its goal state on data from both robot and human activities. By labeling latent actions, IGOR facilitates the learning of foundation policy and world models from internet-scale human video data, covering a diverse range of embodied AI tasks. With a semantically consistent latent action space, IGOR enables human-to-robot generalization. The foundation policy model acts as a high-level controller at the latent action level, which is then integrated with a low-level policy to achieve effective robot control.
    }
   \label{fig:demo1}
\end{figure*}

\pagestyle{fancy}
\fancyhf{}
\fancyhead[C]{\textbf{IGOR}: \textbf{I}mage-\textbf{GO}al \textbf{R}epresentations}
\renewcommand{\footrulewidth}{0.1pt}
\fancyfoot[R]{\thepage\quad}

\fancyfootnotetext{\textit{Project Lead: Li Zhao (\hyperlink{lizo@microsoft.com}{lizo@microsoft.com})}}

\section{Introduction}

Learning foundation models for embodied AI has been notably constrained by a lack of interaction data. 
Unlike text or video data, which are abundantly available, interaction data is much scarcer. Research efforts have been devoted to creating large-scale interaction datasets, such as Open-X-Embodiment~\citep{open_x_embodiment_rt_x_2023} and DROID~\citep{khazatsky2024droid}. 
Based on multi-task interaction data, a series of generalist agents~(or foundation policy models) have been proposed, such as RT-1~\citep{brohan2022rt}, Robocat~\citep{Robocat}, RT-2~\citep{rt22023arxiv}, Octo~\citep{team2024octo}, and OpenVLA~\citep{kim2024openvla}. 
However, the volume of interaction data remains several orders of magnitude smaller than that of internet text or video data. 
Given that the success of foundation models relies on scaling up datasets and extracting knowledge from such large-scale datasets, it is essential to design methods for building embodied AI foundation models that can effectively utilize internet-scale video data.

Internet-scale video data contains abundant sequential records of human activities and perfect demonstrations of how human perform various tasks by interacting with the real world. When the human brain extracts information from videos, instead of doing it frame by frame, it modularizes the differences between frames into a single word such as ``move", ``open", ``close". We refer to these highly compressed, modularized actions as latent actions 
that are shared across different tasks. The question to ask here is, \textbf{is it possible to recover latent actions from video datasets with humans and robots performing various real embodied AI tasks?} While recent works such as Genie~
\citep{bruce2024genie} and LAPO~\citep{schmidt2023learning} made attempts in recovering such latent actions from videos, they primarily focus on 2D platformer games where each latent action $a_t$ corresponds to a specific control button. The action space is highly designed to fit a specific scenario and incomparable to the complex human and robot action space in various embodied AI tasks. To take a step further, the question would be, \textbf{can we learn a unified, semantically consistent latent action space, allowing the transfer of knowledge across different tasks, and embodiments including human and various robots?}

In this paper, we propose Image-GOal Representations (IGOR), which learns a unified and semantically consistent latent action space shared across different tasks and embodiments, enabling the knowledge transfer among internet-scale video data. 
We propose a latent action model designed to capture robot and human actions across various embodied AI tasks. IGOR compresses the visual changes between an image and its goal state into latent actions, which are also embeddings of sub-tasks defined by reaching the goal from the initial image.
IGOR is trained by minimizing the reconstruction loss of the goal state, which is predicted based on the image and the latent action. 
The core insight behind IGOR is that if compressed sufficiently, the image-goal pairs with similar visual changes will have similar embeddings. 

\note{\textit{We argue that, besides text embeddings for human instruction understanding and image/video embeddings for state understanding, image-goal representations for latent action learning and sub-task understanding are yet another crucial building blocks, which may hold great potential for next-level generalization in embodied AI. }}

With the latent action model, we can transform internet-scale human video data into interaction data labeled with latent actions, which largely expands the data available to building embodied AI foundation models.
This unified latent action space allows us to train foundation policy and world models on nearly arbitrary tasks performed by robots and humans.
Specifically, we train a foundation policy model on large-scale video data with text labels. This model uses text to describe tasks and makes decisions, generating the next latent action to perform. 
Additionally, we train a foundation world model on the same dataset, learning to simulate the outcome of executing the foundation policy model.
Image-Goal Representations can be viewed as atomic control units in visual space. They function both as latent actions for a foundational policy model to predict in visual trajectory planning and as sub-tasks for a robot-specific low-level policy to execute.

\begin{figure}[tb!]
    \centering
    \includegraphics[width=0.96\textwidth]{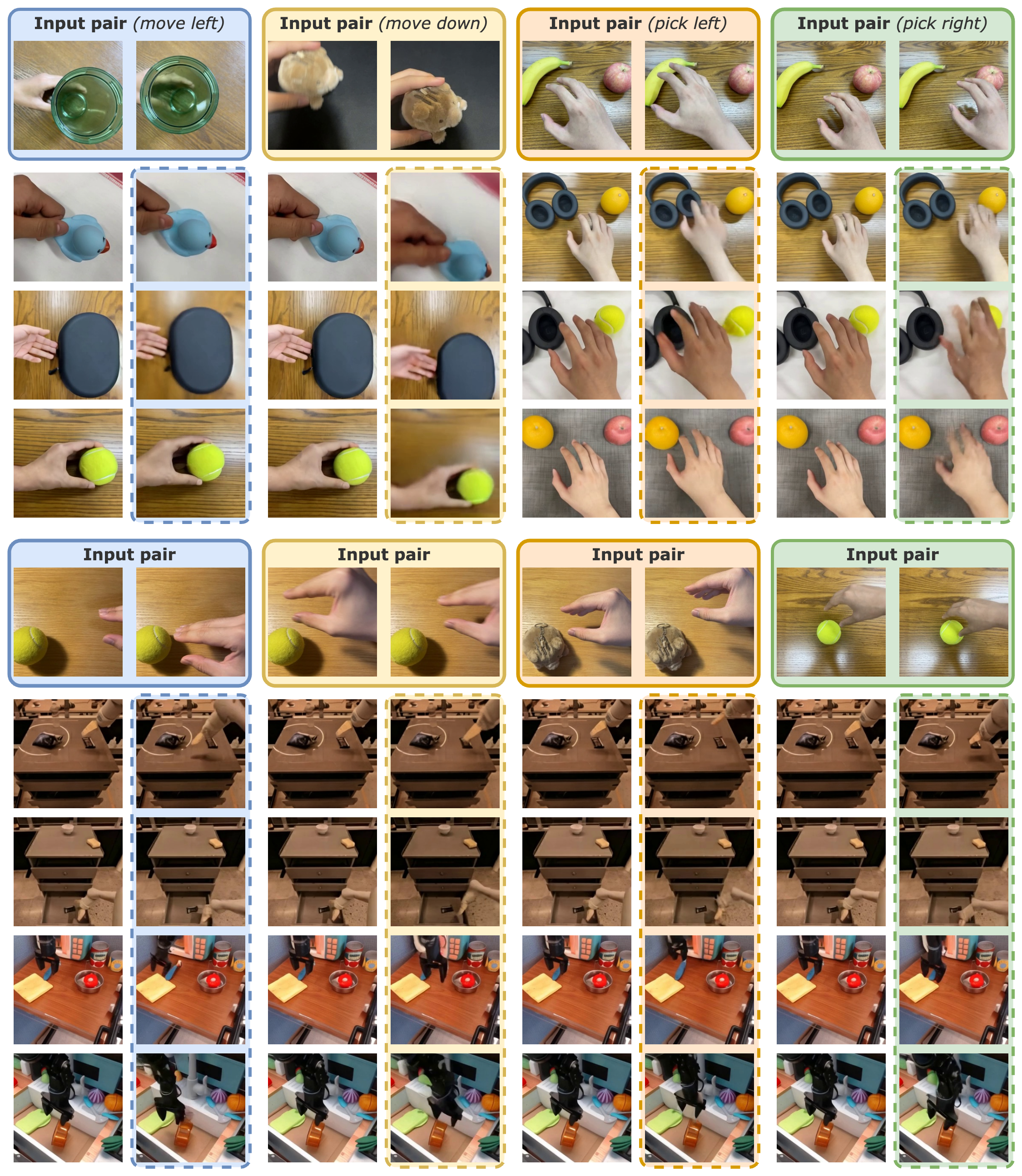}
    \caption{ We extract latent actions from Image-Goal pairs in the solid line boxes, and apply the latent actions to different initial frames, generating subsequent videos via world model as shown in the corresponding dashed boxes. 
    The first half illustrates examples from real-world videos with diverse object categories, while the second half demonstrates generalization from human to robot arms. Full videos are available on our \href{https://aka.ms/project-igor}{website}.
    }
   \label{fig:apply_embedding_real_world}
\end{figure}

We train our models on human video data and robot data with actions removed, with RT-1 dataset held out for OOD evaluation. First, we evaluate the latent action model qualitatively, and find that image-goal pairs with similar latent actions have similar visual changes, corresponding to semantically consistent movements, even on OOD scenarios. Then we evaluate the world model by extracting latent actions from a video and applying such latent action (or action sequence) to 
the initial frames of other videos, generating the rest of frames.
We find that, jointly with the latent action model and world model, IGOR successfully ``migrates'' the movements of the object in the one video to other videos, as shown in Figure~\ref{fig:apply_embedding_real_world}. 
We also apply different latent actions to the same initial image, and find that the world model has learned various possible movements of the object in the image, suggesting that it has absorbed the physical interaction knowledge.
For the foundation policy model, 
we show its ability in following diverse language instruction via iteratively rolling out the foundation policy and world model using latent actions.
We further integrate it with a low-level policy, and show that IGOR-based policy training can improve performance on Google Robot tasks in low-data regime with the SIMPLER~\citep{li24simpler} simulator.

\section{Methodology}

\subsection{Latent Action Model}

The primary objective of the latent action model 
is to label latent actions from unlabeled open-domain videos in an unsupervised manner. Given a sequence of video frames $o_{1:t+1}$, the goal is to derive the latent action $a_t$, which captures the key information describing only the changes that occur at time step $t$, removing other redundant information.
In contrast to prior works \citep{schmidt2023learning, bruce2024genie}, which primarily focus on 2D platformer games where each latent action $a_t$ corresponds to a specific control button, we aim to develop a more generalizable model.
Our model is designed to handle the significantly greater complexity of open-world scenarios, where latent actions may not correspond to any specific underlying actions.
This presents several additional challenges.

First, rather than focusing solely on absolute position of pixel changes, the latent action model must learn to capture semantic movements that remain consistent across varying scenarios. Moreover, due to the temporal redundancy, actions are often sparse given long contexts, which can lead the model to infer $o_{t+1}$ directly from the history, bypassing the need for a more informative latent action $a_t$.

To address these issues, we propose a novel model architecture.
Our latent action model consists of a pair of Inverse Dynamics Model (IDM) and Forward Dynamics Model (FDM).
IDM $I$ is trained to predict the latent action $a_t$ based on the full sequence of observations $o_{1:t+1}$. 
Instead of using the raw observations, we first apply random cropping $c_1$ to the inputs: $a_t = I \left( c_1[o_{1:t+1}] \right)$.
For the architecture of $I$, we first extract features for each frame through Vision Transformer (ViT) \citep{dosovitskiy2021ViT} and then adopt a Spatio-Temporal transformer (ST-transformer) \citep{bruce2024genie, xu2021STtransformer} with a temporal causal mask as the backbone. 
Learnable readout tokens are then used to extract and compress the 
visual changes
into $N$ tokens. To further compress the information stored in latent action, we apply vector quantization to each token, restricting them to a discrete codebook of size $|C|$. Finally, we derive the latent action $a_t \in \mathbf{R}^{N\times D}$ where $D$ is the dimension of each code. We refer to $a_t$ as the latent action embedding, or sub-task embedding, as they describe the information that takes the observation $o_t$ to the next observation $o_{t+1}$. 

For the FDM $F$, 
we propose using a single-frame Vision Transformer to reconstruct $o_{t+1}$, in contrast to previous works \citep{schmidt2023learning, bruce2024genie}, which reconstruct the next frame given the entire context $o_{1:t}$.
This approach mitigates the case where the model might predict the next frame directly from the context, bypassing the latent action. By conditioning on a single frame, it encourages more information to flow into the latent action $a_t$.
For reconstruction, we apply another random cropping $c_2$, and the next frame is predicted as $\hat{o}_{t+1} = F\left( c_2[o_t], a_t \right)$. By using different croppings $c_1$ and $c_2$, the model is encouraged to learn a more semantically invariant latent action across different trajectories.
The models are trained jointly with the reconstruction loss $\| c_2[o_{t+1}] - \hat{o}_{t+1} \|^2$ and the commitment loss in vector quantization.

\subsection{Foundation World Model}

Our foundation world model is a continuous-time Rectified Flow~\citep{liu2022flow,esser2024scaling} that learns to predict the future frames $o_{t+1:T}$ conditioned on the history observation frame $o_{1:t}$, and future latent actions $a_{t:T-1}$. To achieve this goal, there are two key challenges: 1) Generating the photo-realistic frame that describes the states precisely; 2) Controlling the generated frames by the latent actions.

Accordingly, we start our foundation world model with the pre-trained Open-Sora~\citep{opensora}. It consists of two components: a 3D Variational AutoEncoder (VAE) that encodes the raw observation into latent space with $8\times8$ times downsampling in spatial dimension and $4\times$ times downsampling in temporal dimension; a Spatial-Temporal Rectified Flow Transformer~(ST-RFT) that generates the latent from the text conditions. To enable the control from the observation and action, we make two modifications to the original Open-Sora: 1) We replace the original text input of the pre-trained model with our latent actions $a_{1:T}$ obtained from LAM.
Zero-padding is applied for the last action.
For each frame, we map the latent actions into a single token and feed it to the ST-RFT via the cross-attention mechanism; 2) We also make the generation conditioned on the output of FDM $\hat{o}_{t+1:T}$, which provides a coarse-grained prediction according to the input latent action. For the conditioning of $\hat{o}_{t+1:T}$, we encode it to the latent space with the same 3D VAE and directly add it to the noisy input element-wise.

Formally, Rectified Flow~\citep{liu2022flow,albergobuilding,esser2024scaling} aims at directly regressing a vector field that generates a probability path between noise distribution and data distribution. For $n \in [0, 1]$, we define the interpolation between the two distributions as:
\begin{equation}
\boldsymbol{x}_n = (1-n) \boldsymbol{x}_0 + n \boldsymbol{x}_1,
\end{equation}
where $x_0$ is the clean data, $x_1$ is the sampled noise, and $x_n$ is the noisy data. During training, we train a vector-valued neural network $\boldsymbol{x}_\theta$ with L2 loss:
\begin{equation}
\mathbb{E}_{n, \boldsymbol{x}_0, \boldsymbol{x}_1} { \left\| \boldsymbol{x}_0 - \boldsymbol{x}_\theta(\boldsymbol{x}_n, n, a_{t:T-1}, \hat{o}_{t+1:T}) \right\|^2}.
\end{equation}
Instead of predicting the conditional expectation directly, we follow~\cite{liu2022flow} to parameterize the velocity with a neural network $\boldsymbol{v}_\theta$ and train it on:
\begin{equation}
L_\mathrm{world}(\theta) = \mathbb{E}_{n, \boldsymbol{x}_0, \boldsymbol{x}_1} { \left\| (\boldsymbol{x}_1 - \boldsymbol{x}_0) - \boldsymbol{v}_\theta(\boldsymbol{x}_n, n, a_{t:T-1}, \hat{o}_{t+1:T}) \right\|^2}.
\end{equation}

It should be noted that, our foundation world model can be fine-tuned to accommodate the different action spaces of robots with various embodiments. The fine-tuning of the foundational world model is left as future work.

\subsection{Foundation Policy Model and Low-level Policy Model}

The training of the policy model consists of two stages. In the first pretraining stage, taken as input the raw observation frames $o_{1:t}$ and a textual description $s$ for the task, the foundation policy model predicts latent actions $a_t = I([o_{1:t+1}])$ labeled by the IDM in the latent action model at each step. The training dataset of this stage is the same as that used for the latent action model, i.e., with large-scale and diverse sources of videos.
In the second finetuning stage, we add an extra prediction component on the foundation policy model to predict real continuous robot actions, with taking the raw observations as well as the latent actions predicted by the first stage model as input. In this stage, only the prediction component~(i.e., the low-level policy model) is optimized on small-scale and task-specific downstream datasets, while other components are frozen.

Specifically, similar to the latent action model, the backbone of foundation policy model is also a ST-transformer equipped with a ViT image encoder, with a feed-forward layer as the final prediction layer.
The textual description $s$ is encoded to a latent representation by a pre-trained text encoder, which is concatenated with the observation representation encoded by the ViT encoder as the joint input to the model.

We use the L2 distance between the predicted hidden output and the latent action as the loss function. Given a trajectory consists of $t$ observations $o_{1:t}$, the training objective can be written as: 
\begin{equation}
    L_{policy} = \|  P([s;o_{1:t} ]) - a_{t} \|^2,
    \label{equ:policy-pretrain}
\end{equation}
where $P(\cdot)$ denotes the policy model.

In the second stage, we train the low-level policy model to predict the real continuous actions within each latent action, where the image-goal latent actions can be seen as representations for sub-tasks defined by reaching a goal from an initial image.
The low-level policy model is also an ST-transformer with a prediction layer. The input consists of the textual representation $s$, the observation $o_{1:t}$ and latent actions predicted by the foundation policy model $P([s;o_{1:t}])$, which are concatenated together at the patch level as one part. The latent action $P([s;o_{1:t}])$ predicted by the foundation policy model also serves as sub-task embedding for the low-level policy model. We denote that each latent action corresponds to $\tau$ real robot actions, and the latent action $a_t$ corresponds to real robot action $u_{t}^{1:\tau}$. Denote the low-level policy model as $P_f(\cdot)$, we train the second stage model also by L2 distance:
\begin{equation}
    L_{ft} = \| P_f([s; P([s; o_{1:t}]); o_{1:t}]) - u_{t}^{1:\tau} \|^2,
    \label{equ:policy-pretrain-2}
\end{equation}
where only the parameters of the low-level policy are optimized.

\section{Experiments}

\subsection{Dataset}

In the pretraining stage, we construct a large-scale dataset comprising diverse domains, including robotic data from various embodiments and a substantial amount of human activity videos. 

\paragraph{Data Mixture.} For the robotic data, we select a subset of Open-X Embodiment dataset \citep{open_x_embodiment_rt_x_2023} with single arm end-effector control, excluding RT-1 dataset for out-of-distribution (OOD) evaluation.
We follow the preprocessing and data mixture weights from \cite{team2024octo, kim2024openvla}.
In total, we utilize approximately 0.8M robot trajectories. 
While our dataset includes data from real robots, we discard the associated actions and proprio-states, using only image frames and text instructions during pretraining.
Additionally, we incorporate large-scale open-world videos with language instructions, including human daily activities from Something-Something v2 \citep{goyal2017something}, and egocentric videos such as EGTEA \citep{li2018eye}, Epic Kitchen \citep{damen2020epic}, and Ego4D \citep{grauman2022ego4d, pramanick2023egovlpv2}. 
In total, we derive approximately 2.0M human activity video clips with high quality. 
Overall, our dataset for pretraining comprises around 2.8M trajectories and video clips, where each trajectory contains a language instruction and a sequence of observations.

\paragraph{Data Preprocessing. } In practice, we found that the video quality has a big impact on the model performance.
We exclude low-quality videos characterized by excessive shakiness or rapid camera movement, and apply stabilization techniques to the remaining videos. 
To ensure proper amount of changes between frames in the latent action model, we choose the optimal frame rates for robotics dataset and human activity videos.

In the finetuning stage, we use RT-1 dataset, a large-scale dataset for real-world robotic experiences. We uniformly sample 1\% number of episodes from RT-1 dataset for finetuning, where each episode comprises a language instruction, a sequence of image observations, and a sequence of low-level actions. The action space is 7-dimensional, including 3 dimensions of robot arm movement $\Delta Pos$, 3 dimensions of robot arm rotation $\Delta Rot$, and 1 dimension of robot gripper action $\Delta Grp$.   We provide more details in Appendix \ref{appendix:dataset}.
\subsection{Training Details}

\begin{figure}[t!] 
    \centering
    \includegraphics[width=1.0\textwidth]{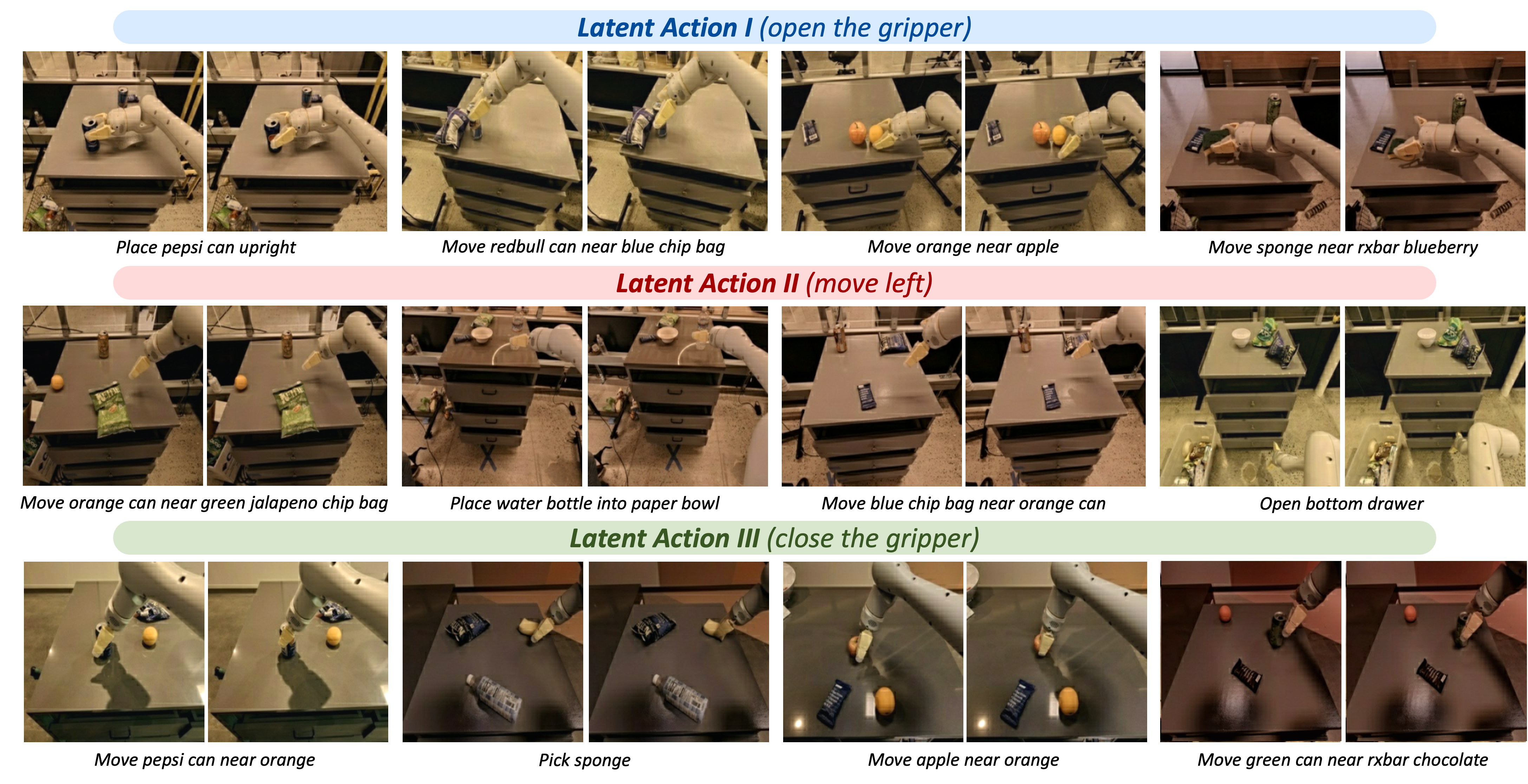}
    \caption{Image-goal pairs with similar latent actions
    in OOD RT-1 dataset.
    In each row, we choose the leftmost image-goal pair, and retrieve 3 nearest pairs on latent action embedding. The original task instructions of the pairs are shown under the images. We find that each row shares the similar visual changes semantically, and the latent actions generalize across different raw language tasks.
    }
   \label{fig:sim_embeddings1}
\end{figure}

We first pretrain our latent action model on our pretraining dataset. Then, we use the pretrained latent action model to label latent actions on our pretraining dataset, and pretrain foundation policy model and foundation world model on the labeled dataset. Finally, we finetune our low-level policy model on top of our pretrained models on RT-1. 

For latent action model, we use a codebook with $N=4$ tokens, and codebook size of $|C|=32$, each with an embedding size of $D=128$. We use a sub-task length of $\tau=3$ for finetuning the low-level policy model on RT-1 dataset. Please refer to Appendix \ref{appendix:training_details} for more training details.

\subsection{Qualitative Results on Latent Actions}

\begin{figure}[tb!]
    \centering
    \includegraphics[width=0.5\textwidth]{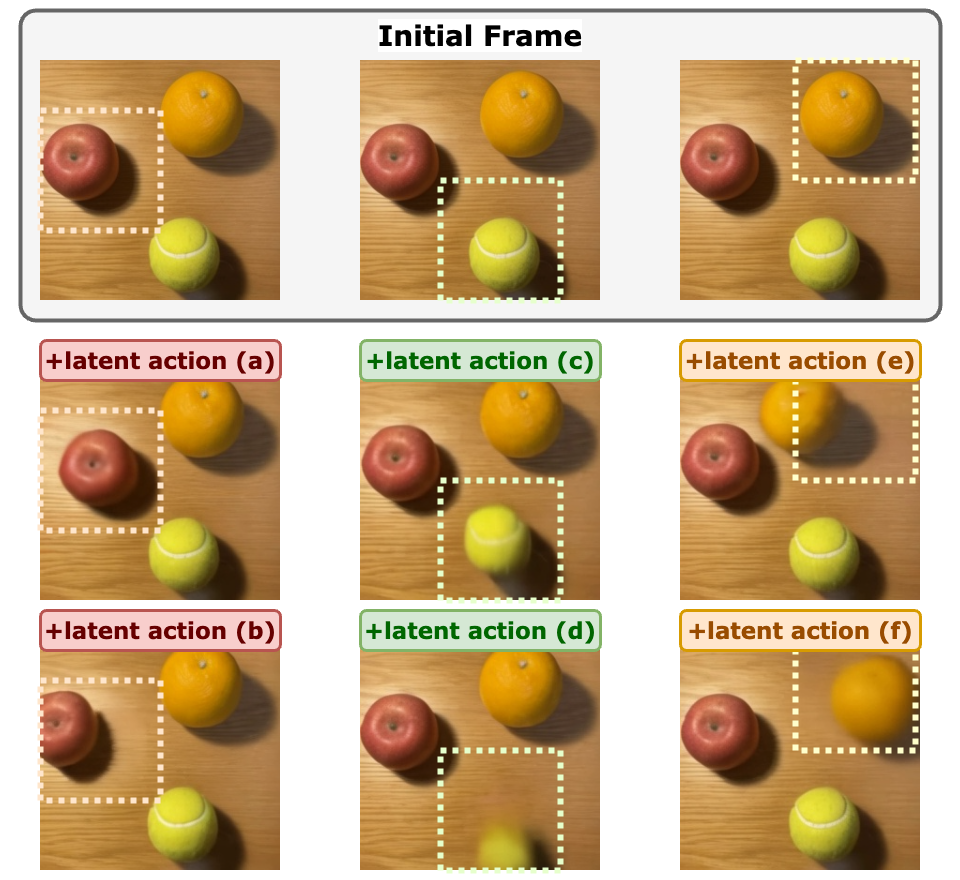}
    \caption{ Controllability of latent action among multiple objects. The last two rows show the generated image by applying 6 different latent actions to the initial frame. Effects of applying different latent actions are highlighted in dashed squares: (a,b) move the apple, (c,d) move the tennis, (e,f) move the orange. Full generated videos from the world model are available on our \href{https://aka.ms/project-igor}{webpage}. 
    }
   \label{fig:controllability}
\end{figure}

We present qualitative results on latent actions learned from robotics and human activity dataset. Specifically, we answer the following questions on learned latent actions:

\begin{itemize}
    \item Do similar latent actions reflect similar visual changes? 
    \item Can latent actions encode semantically consistent changes across different tasks, and embodiments including human and robots? If so, are we able to migrate movements in videos across embodiments and tasks via latent action? 
    \item Does the policy foundation model properly follow language instructions for task solving? 
\end{itemize}

\subsubsection{
Visualization of Image-Goal Pairs with Similar Latent Actions 
}

We investigate whether similar learned latent actions reflect similar visual changes on robotics manipulation dataset. 
We use RT-1 dataset, which was excluded from the latent action model training and serves as out-of-domain samples for evaluation.
We randomly select image-goal pairs from RT-1 dataset, and present the image-goal pairs with smallest euclidean distance in latent action embedding in RT-1 dataset 
in Figure~\ref{fig:sim_embeddings1}. We observe that pairs with similar embeddings indeed have similar visual changes, and also similar sub-tasks in semantic, for example, ``open the gripper'', ``move left'', and ``close the gripper''.
Furthermore, each sub-task appears in different raw language tasks, suggesting the latent actions are reused, thereby facilitating generalization in model learning.

\subsubsection{Controllability of Latent Actions}
\label{sec:controllability}

We demonstrate that latent actions are able to control the changes in objects on different real world scenes, and the effects of latent actions generalize across tasks and embodiments. 
Specially, the generalizability of latent actions enable IGOR to successfully migrate human movement videos into robot movements provided the initial image, despite they largely differ in embodiments.

\paragraph{Object Controllability Among Multiple Objects. }  
We evaluate the controllability of the latent actions on object movements among multiple objects on the same image. 
In Figure~\ref{fig:controllability}, we generate subsequent images by applying 6 different actions to the same original image on the foundation world model. We observe that the latent action model and foundation world model learn to control specific object's movement among multiple objects.

\paragraph{Object Controllability Across Embodiments and Tasks.} 
We evaluate the semantic consistency of the latent actions across different setups, including embodiments and tasks. 
We use pairs of image-goal in the real world manipulation videos to generate latent actions, and apply the same set of actions to other images in different scene setups with foundation world model to generate subsequent videos. 
The results are shown in Figure~\ref{fig:apply_embedding_real_world}. Impressively, we observe that (1) latent actions are semantically consistent across different tasks with different object categories; (2) latent actions are semantically consistent across human and robots. By applying latent actions extracted from human demonstrations, we generate videos of robot arm movements.
With only one demonstration, the robot arm can successfully migrate human behaviors, which opens up new possibilities for few-shot human-to-robot transfer and control.

\subsubsection{Counterfactual Video Generation with Diverse Instructions}

\begin{figure}[t!]
    \centering
    \includegraphics[width=1.0\textwidth]{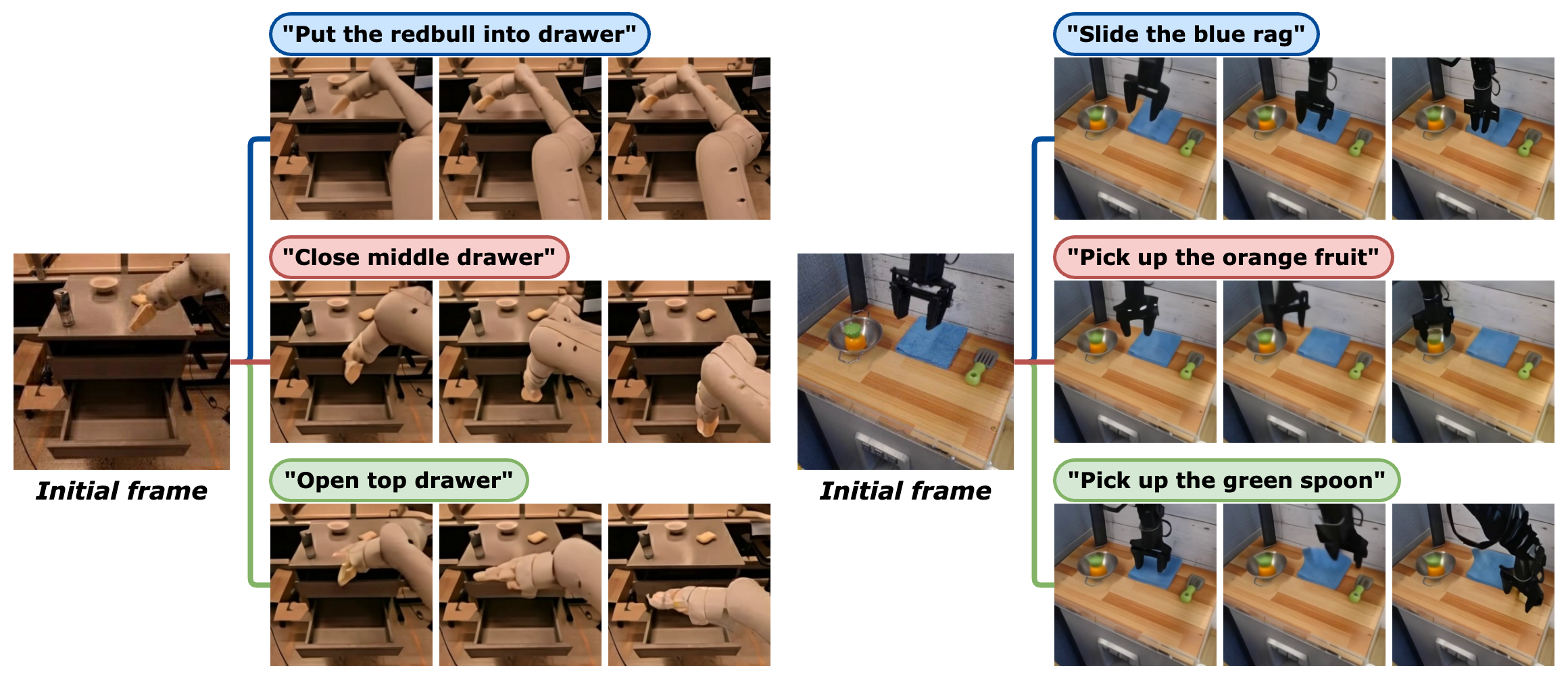}
    \caption{Generated image sequence jointly by the foundation policy and world model via only latent actions, following 3 different instructions from the same initial image. Full generated videos from the world model are available on our \href{https://aka.ms/project-igor}{webpage}.
    }
   \label{fig:counterfactual2}
\end{figure}

We analyze whether the foundation policy model has the ability to follow human instructions. To this end, we interpret the effect of latent actions visually with the foundation world model. Starting from a single initial image, the foundation policy and world model can jointly generate diverse behaviors in videos that follow diverse instructions using only latent actions. We experiment with initial images from RT-1 and Bridge dataset and manually written instructions, and show the image clips of generated videos in Figure~\ref{fig:counterfactual2}. The results show that the foundation policy model can properly follow different language instructions for task solving. 

\subsection{Quantitative Results}
 \subsubsection{Evaluation on the Google Robot Tasks in SIMPLER}

We evaluate our IGOR-based training framework on the Google robot tasks in the SIMPLER simulator under a low-data regime, utilizing only 1\% of the data from the large RT-1 dataset for the low-level policy learning stage. 
\paragraph{Evaluation Setups. } We test different model’s ability to control the Google Robot following language tasks with RGB images as observations, where all robots are controlled with low-level end-effector control actions, after finetuning on the same amount of data from RT-1 dataset. We evaluate the success rate on three tasks: ``Pick Coke Can'', ``Move Near'', and ``Open / Close Drawer''. 

\paragraph{Baseline Method. } We compare our method with the same low-level policy model architecture with ST-Transformer, without latent action embedding concatenated on the observation feature embedding.

\begin{figure}[t]
    \centering
    \includegraphics[width=1.0\textwidth]{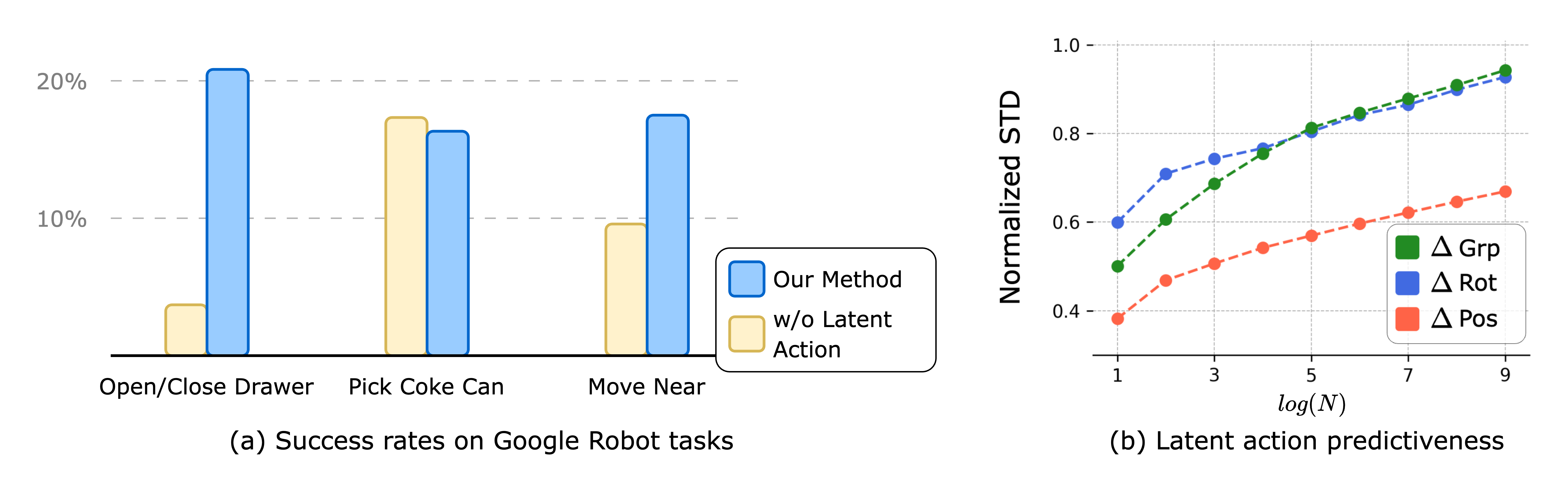}
    \caption{ \textbf{(a).} Success rate of IGOR and the low-level policy trained from scratch methods on Google Robot tasks under SIMPLER simulator, finetuned on 1\% data of RT-1. 
    \textbf{(b).} Predictiveness of latent action on robot action. X-axis: $\log(N)$, where $N$ is the number of nearest neighbours in latent action embedding. Y-axis: normalized standard deviation in action embedding with respect to movement actions (orange), rotation actions (blue), and gripper actions (green).}
   \label{fig:simplerandstd}
\end{figure}

We present the success rate of different methods in Figure~\ref{fig:simplerandstd}(a).
From the figure, we observe that IGOR achieves higher or equal success rate than the model trained from scratch, showing the generalizability of the learned latent action to real robotics actions. 

\subsubsection{Predictiveness of Latent Actions on Robot Actions}

We analyze whether our learned latent actions are predictive of real robot actions. On RT-1 dataset, we randomly sample a number of $M=15,000$ pairs of images, and compute their latent action embeddings. For each pair of image, we find $N$ nearest neighbours of image pairs in RT-1 dataset with the closest latent action embedding, and compute the standard deviation of real robot actions among $N$ neighbours
on each action dimension, normalized by the standard deviation of robot actions over each dimension over the whole RT-1 dataset.
By varying $N$, we assess whether closer latent actions correspond to more similar downstream actions.

The results are shown in Figure~\ref{fig:simplerandstd}(b). 
The fact that smaller $N$ leading to lower normalized standard deviation, and all normalized standard deviation being below $1.0$,
show that the latent actions are predictive of real robot actions including robot movements, rotations and gripper actions.
It is also shown that the latent actions are more predictive of the robot movement than rotations and gripper actions, suggesting that the IGOR learned action space reflects more information in robot movements than robot arm rotations and gripping. 
\subsection{Ablation Studies}

We provide additional ablation studies on the pretraining dataset 
of latent action model, showing that using a mixture of robotics and human activity dataset benefits the generalization of latent action model.
Detailed ablation studies results are provided in Appendix~\ref{appendix:ablation}. 

\section{Related Work}

\paragraph{Foundation Agents for Robots}
Open-ended task-agnostic training and high-capacity neural network architectures have been recognized as key to the success of foundation models. In this context, a series of generalist agents have been proposed as the foundation policy models for robots~\citep{brohan2022rt, Robocat,rt22023arxiv,team2024octo,kim2024openvla}. RT-1~\citep{brohan2022rt} contributes a large-scale multi-task dataset and a robotic transformer architecture,facilitating and assessing generalization across multiple tasks. RoboCat builds on Gato~\citep{reed2022generalistagent}, further enabling multi-embodiment generalization.  RT-2 highlights the importance of leveraging vision-language models trained on internet-scale data~\citep{rt22023arxiv}.  Octo~\citep{team2024octo} and OpenVLA~\citep{kim2024openvla} can be seen as open versions of RoboCat and RT-2 respectively, with some additional technical contributions. IGOR is similar to RT-2 and OpenVLA in the sense that we both leverage Internet-scale data. The difference lies in that we use video data~(with text labels) of human/robot performing embodied AI tasks, while they use text data and visual question answering data for the training of vision language models. To the best of our knowledge, we present the first foundation policy model that performs decision making at the sub-task (i.e. latent action) level. 

\paragraph{Image-Goal Visual Changes}
Tracking visual changes and establishing correspondence between an image and its goal state is crucial for dynamic visual understanding in embodied AI. SiamMAE~\citep{gupta2023siamMAE} proposes to use a siamese encoder on the image and goal to learn visual correspondence. Voltron~\citep{karamcheti2023voltron} introduces language-guided visual representation learning on image-goal pairs. FLOWRETRIEVAL~\citep{lin2024flowretrieval} and AVDC~\citep{ko2023learningactactionlessvideos} leverage optical flow between image and goal to capture visual changes and correspondence, while Video-LaVIT~\citep{jin2024videolavit} utilizes motion vectors. 
iVideoGPT~\citep{wu2024ivideogpt} proposes using image-conditioned goal representations as state representations to predict within a world model.
VPT~\citep{baker2022videopretraining} proposes to recover latent actions in videos using an inverse dynamics model trained on interaction data to predict real actions.
Perhaps the most similar approaches to our methods are LAPO~\citep{schmidt2023learning} and Genie~\citep{bruce2024genie}. Both works primarily focus on 2D platformer games where each latent action corresponds to a specific control button. By contrast, we aim to develop a more generalizable model to handle the significantly greater complexity of open-world scenarios, where latent actions may not correspond to any specific underlying actions. 

\paragraph{Video Generation for Embodied AI}

Video generation is another research topic closely related to embodied AI. It has been proposed that video can be seen as the new language for real-world decision making~\citep{yang2024video}.
Many works on world models build on video generation techniques~\citep{bruce2024genie, wu2024ivideogpt,hu2023gaia1generativeworldmodel,yang2024UniSim, xiang2024pandorageneralworldmodel}. Some text-to-video works claim to be real-world simulators, such as Sora~\citep{videoworldsimulators2024} and WorldDreamer~\citep{wang2024worlddreamer}. 
Unipi~\citep{du2023unipi} proposes to first predict the next goal state, then infer real robot actions with an inverse dynamics model. 
By contrast, our foundation policy model first predicts the latent action, which can specify the goal state, and then uses the latent action to enable sub-task level generalization. We argue that forward prediction in latent action space, rather than the original image space, offers several advantages. For example, we can perform sub-task understanding for image-goal representations, and the compressed latent action could be easier to predict than the entire image. 

\paragraph{Pre-trained Visual Representations}

Pre-trained Visual Representations target on training representations for images/videos in self-supervised learning manner~\citep{he2021MAE, Xiao2022, Radosavovic2022, vc2023, radford2021CLIP, nair2022r3m, ma2023vip, oquab2023dinov2, darcet2023vitneedreg, kirillov2023segment,assran2023JEPA,  bardes2024VJEPA}, 
and has been demonstrated to be very effective for state understanding in embodied AI.
By contrast, IGOR learns image-goal representations for sub-task understanding, which we believe are another crucial building blocks, that may significantly enhance generalization in embodied AI.

\section{Conclusions, Limitations, and Future Work}

In this paper, we propose IGOR, a novel training framework, taking the first step towards learning a unified action space for humans and robots in various embodied AI tasks. 

Qualitatively, we demonstrate that: 

\begin{itemize}
    \item IGOR learns similar representations for image pairs with similar visual changes.
    \item The learned latent action has control over the next state given the current image.
    \item The foundation world model acquires knowledge about objects and their potential movements.
    \item The foundation policy model learns to follow instructions across different states.
\end{itemize}

Quantitatively, we show that: 

\begin{itemize}
    \item On RT-1 dataset, image-goal pairs with similar latent actions are associated with similar low-level robot actions.
    \item The IGOR framework improves policy learning, potentially due to its capability to predict the next sub-task by leveraging internet-scale data, thereby enabling sub-task level generalization.
\end{itemize}

The IGOR framework is limited in the following perspective: we cannot separate visual changes caused by the agents, other agents~(such as dogs), or the  shakiness of camera. To address this, we mitigated camera shakiness and used only ego-centric videos without other agents in view. Just like any other representation learning methods, scaling up the dataset and model size is always most straightforward and effective. To facilitate the usage of more data, incorporating image processing methods such as object segmentation with IGOR will be part of future works. For better applications in embodied AI, the foundation world model can also be tuned to match real world scenarios, along with other improvements such as adapting the latent action model for multi-agent scenarios.

\bibliography{main}

\begin{thebibliography}{75}
\providecommand{\natexlab}[1]{#1}
\providecommand{\url}[1]{\texttt{#1}}
\expandafter\ifx\csname urlstyle\endcsname\relax
  \providecommand{\doi}[1]{doi: #1}\else
  \providecommand{\doi}{doi: \begingroup \urlstyle{rm}\Url}\fi

\bibitem[Albergo \& Vanden-Eijnden(2023)Albergo and Vanden-Eijnden]{albergobuilding}
Albergo, M.~S. and Vanden-Eijnden, E.
\newblock Building normalizing flows with stochastic interpolants.
\newblock In \emph{International Conference on Learning Representations}, 2023.

\bibitem[Assran et~al.(2023)Assran, Duval, Misra, Bojanowski, Vincent, Rabbat, LeCun, and Ballas]{assran2023JEPA}
Assran, M., Duval, Q., Misra, I., Bojanowski, P., Vincent, P., Rabbat, M., LeCun, Y., and Ballas, N.
\newblock Self-supervised learning from images with a joint-embedding predictive architecture, 2023.
\newblock URL \url{https://arxiv.org/abs/2301.08243}.

\bibitem[Baker et~al.(2022)Baker, Akkaya, Zhokhov, Huizinga, Tang, Ecoffet, Houghton, Sampedro, and Clune]{baker2022videopretraining}
Baker, B., Akkaya, I., Zhokhov, P., Huizinga, J., Tang, J., Ecoffet, A., Houghton, B., Sampedro, R., and Clune, J.
\newblock Video pretraining (vpt): Learning to act by watching unlabeled online videos, 2022.
\newblock URL \url{https://arxiv.org/abs/2206.11795}.

\bibitem[Bardes et~al.(2024)Bardes, Garrido, Ponce, Chen, Rabbat, LeCun, Assran, and Ballas]{bardes2024VJEPA}
Bardes, A., Garrido, Q., Ponce, J., Chen, X., Rabbat, M., LeCun, Y., Assran, M., and Ballas, N.
\newblock Revisiting feature prediction for learning visual representations from video, 2024.
\newblock URL \url{https://arxiv.org/abs/2404.08471}.

\bibitem[Belkhale et~al.(2023)Belkhale, Cui, and Sadigh]{belkhale2023hydra}
Belkhale, S., Cui, Y., and Sadigh, D.
\newblock Hydra: Hybrid robot actions for imitation learning.
\newblock \emph{arxiv}, 2023.

\bibitem[Bousmalis et~al.(2023)Bousmalis, Vezzani, Rao, Devin, Lee, Bauza, Davchev, Zhou, Gupta, Raju, Laurens, Fantacci, Dalibard, Zambelli, Martins, Pevceviciute, Blokzijl, Denil, Batchelor, Lampe, Parisotto, Żołna, Reed, Colmenarejo, Scholz, Abdolmaleki, Groth, Regli, Sushkov, Rothörl, Chen, Aytar, Barker, Ortiz, Riedmiller, Springenberg, Hadsell, Nori, and Heess]{Robocat}
Bousmalis, K., Vezzani, G., Rao, D., Devin, C., Lee, A.~X., Bauza, M., Davchev, T., Zhou, Y., Gupta, A., Raju, A., Laurens, A., Fantacci, C., Dalibard, V., Zambelli, M., Martins, M., Pevceviciute, R., Blokzijl, M., Denil, M., Batchelor, N., Lampe, T., Parisotto, E., Żołna, K., Reed, S., Colmenarejo, S.~G., Scholz, J., Abdolmaleki, A., Groth, O., Regli, J.-B., Sushkov, O., Rothörl, T., Chen, J.~E., Aytar, Y., Barker, D., Ortiz, J., Riedmiller, M., Springenberg, J.~T., Hadsell, R., Nori, F., and Heess, N.
\newblock Robocat: A self-improving generalist agent for robotic manipulation, 2023.
\newblock URL \url{https://arxiv.org/abs/2306.11706}.

\bibitem[Brohan et~al.(2022)Brohan, Brown, Carbajal, Chebotar, Dabis, Finn, Gopalakrishnan, Hausman, Herzog, Hsu, et~al.]{brohan2022rt}
Brohan, A., Brown, N., Carbajal, J., Chebotar, Y., Dabis, J., Finn, C., Gopalakrishnan, K., Hausman, K., Herzog, A., Hsu, J., et~al.
\newblock Rt-1: Robotics transformer for real-world control at scale.
\newblock \emph{arXiv preprint arXiv:2212.06817}, 2022.

\bibitem[Brohan et~al.(2023)Brohan, Brown, Carbajal, Chebotar, Chen, Choromanski, Ding, Driess, Dubey, Finn, Florence, Fu, Arenas, Gopalakrishnan, Han, Hausman, Herzog, Hsu, Ichter, Irpan, Joshi, Julian, Kalashnikov, Kuang, Leal, Lee, Lee, Levine, Lu, Michalewski, Mordatch, Pertsch, Rao, Reymann, Ryoo, Salazar, Sanketi, Sermanet, Singh, Singh, Soricut, Tran, Vanhoucke, Vuong, Wahid, Welker, Wohlhart, Wu, Xia, Xiao, Xu, Xu, Yu, and Zitkovich]{rt22023arxiv}
Brohan, A., Brown, N., Carbajal, J., Chebotar, Y., Chen, X., Choromanski, K., Ding, T., Driess, D., Dubey, A., Finn, C., Florence, P., Fu, C., Arenas, M.~G., Gopalakrishnan, K., Han, K., Hausman, K., Herzog, A., Hsu, J., Ichter, B., Irpan, A., Joshi, N., Julian, R., Kalashnikov, D., Kuang, Y., Leal, I., Lee, L., Lee, T.-W.~E., Levine, S., Lu, Y., Michalewski, H., Mordatch, I., Pertsch, K., Rao, K., Reymann, K., Ryoo, M., Salazar, G., Sanketi, P., Sermanet, P., Singh, J., Singh, A., Soricut, R., Tran, H., Vanhoucke, V., Vuong, Q., Wahid, A., Welker, S., Wohlhart, P., Wu, J., Xia, F., Xiao, T., Xu, P., Xu, S., Yu, T., and Zitkovich, B.
\newblock Rt-2: Vision-language-action models transfer web knowledge to robotic control.
\newblock In \emph{arXiv preprint arXiv:2307.15818}, 2023.

\bibitem[Brooks et~al.(2024)Brooks, Peebles, Holmes, DePue, Guo, Jing, Schnurr, Taylor, Luhman, Luhman, Ng, Wang, and Ramesh]{videoworldsimulators2024}
Brooks, T., Peebles, B., Holmes, C., DePue, W., Guo, Y., Jing, L., Schnurr, D., Taylor, J., Luhman, T., Luhman, E., Ng, C., Wang, R., and Ramesh, A.
\newblock Video generation models as world simulators.
\newblock 2024.
\newblock URL \url{https://openai.com/research/video-generation-models-as-world-simulators}.

\bibitem[Bruce et~al.(2024)Bruce, Dennis, Edwards, Parker-Holder, Shi, Hughes, Lai, Mavalankar, Steigerwald, Apps, et~al.]{bruce2024genie}
Bruce, J., Dennis, M.~D., Edwards, A., Parker-Holder, J., Shi, Y., Hughes, E., Lai, M., Mavalankar, A., Steigerwald, R., Apps, C., et~al.
\newblock Genie: Generative interactive environments.
\newblock In \emph{Forty-first International Conference on Machine Learning}, 2024.

\bibitem[Chen et~al.()Chen, Adebola, and Goldberg]{BerkeleyUR5Website}
Chen, L.~Y., Adebola, S., and Goldberg, K.
\newblock Berkeley {UR5} demonstration dataset.
\newblock \url{https://sites.google.com/view/berkeley-ur5/home}.

\bibitem[Collaboration et~al.(2023)Collaboration, O'Neill, Rehman, Maddukuri, Gupta, Padalkar, Lee, Pooley, Gupta, Mandlekar, Jain, Tung, Bewley, Herzog, Irpan, Khazatsky, Rai, Gupta, Wang, Kolobov, Singh, Garg, Kembhavi, Xie, Brohan, Raffin, Sharma, Yavary, Jain, Balakrishna, Wahid, Burgess-Limerick, Kim, Schölkopf, Wulfe, Ichter, Lu, Xu, Le, Finn, Wang, Xu, Chi, Huang, Chan, Agia, Pan, Fu, Devin, Xu, Morton, Driess, Chen, Pathak, Shah, Büchler, Jayaraman, Kalashnikov, Sadigh, Johns, Foster, Liu, Ceola, Xia, Zhao, Frujeri, Stulp, Zhou, Sukhatme, Salhotra, Yan, Feng, Schiavi, Berseth, Kahn, Yang, Wang, Su, Fang, Shi, Bao, Amor, Christensen, Furuta, Walke, Fang, Ha, Mordatch, Radosavovic, Leal, Liang, Abou-Chakra, Kim, Drake, Peters, Schneider, Hsu, Bohg, Bingham, Wu, Gao, Hu, Wu, Wu, Sun, Luo, Gu, Tan, Oh, Wu, Lu, Yang, Malik, Silvério, Hejna, Booher, Tompson, Yang, Salvador, Lim, Han, Wang, Rao, Pertsch, Hausman, Go, Gopalakrishnan, Goldberg, Byrne, Oslund, Kawaharazuka, Black, Lin, Zhang, Ehsani,
  Lekkala, Ellis, Rana, Srinivasan, Fang, Singh, Zeng, Hatch, Hsu, Itti, Chen, Pinto, Fei-Fei, Tan, Fan, Ott, Lee, Weihs, Chen, Lepert, Memmel, Tomizuka, Itkina, Castro, Spero, Du, Ahn, Yip, Zhang, Ding, Heo, Srirama, Sharma, Kim, Kanazawa, Hansen, Heess, Joshi, Suenderhauf, Liu, Palo, Shafiullah, Mees, Kroemer, Bastani, Sanketi, Miller, Yin, Wohlhart, Xu, Fagan, Mitrano, Sermanet, Abbeel, Sundaresan, Chen, Vuong, Rafailov, Tian, Doshi, Mart{'i}n-Mart{'i}n, Baijal, Scalise, Hendrix, Lin, Qian, Zhang, Mendonca, Shah, Hoque, Julian, Bustamante, Kirmani, Levine, Lin, Moore, Bahl, Dass, Sonawani, Song, Xu, Haldar, Karamcheti, Adebola, Guist, Nasiriany, Schaal, Welker, Tian, Ramamoorthy, Dasari, Belkhale, Park, Nair, Mirchandani, Osa, Gupta, Harada, Matsushima, Xiao, Kollar, Yu, Ding, Davchev, Zhao, Armstrong, Darrell, Chung, Jain, Vanhoucke, Zhan, Zhou, Burgard, Chen, Chen, Wang, Zhu, Geng, Liu, Liangwei, Li, Pang, Lu, Ma, Kim, Chebotar, Zhou, Zhu, Wu, Xu, Wang, Bisk, Dou, Cho, Lee, Cui, Cao, Wu, Tang, Zhu,
  Zhang, Jiang, Li, Li, Iwasawa, Matsuo, Ma, Xu, Cui, Zhang, Fu, and Lin]{open_x_embodiment_rt_x_2023}
Collaboration, O. X.-E., O'Neill, A., Rehman, A., Maddukuri, A., Gupta, A., Padalkar, A., Lee, A., Pooley, A., Gupta, A., Mandlekar, A., Jain, A., Tung, A., Bewley, A., Herzog, A., Irpan, A., Khazatsky, A., Rai, A., Gupta, A., Wang, A., Kolobov, A., Singh, A., Garg, A., Kembhavi, A., Xie, A., Brohan, A., Raffin, A., Sharma, A., Yavary, A., Jain, A., Balakrishna, A., Wahid, A., Burgess-Limerick, B., Kim, B., Schölkopf, B., Wulfe, B., Ichter, B., Lu, C., Xu, C., Le, C., Finn, C., Wang, C., Xu, C., Chi, C., Huang, C., Chan, C., Agia, C., Pan, C., Fu, C., Devin, C., Xu, D., Morton, D., Driess, D., Chen, D., Pathak, D., Shah, D., Büchler, D., Jayaraman, D., Kalashnikov, D., Sadigh, D., Johns, E., Foster, E., Liu, F., Ceola, F., Xia, F., Zhao, F., Frujeri, F.~V., Stulp, F., Zhou, G., Sukhatme, G.~S., Salhotra, G., Yan, G., Feng, G., Schiavi, G., Berseth, G., Kahn, G., Yang, G., Wang, G., Su, H., Fang, H.-S., Shi, H., Bao, H., Amor, H.~B., Christensen, H.~I., Furuta, H., Walke, H., Fang, H., Ha, H., Mordatch, I.,
  Radosavovic, I., Leal, I., Liang, J., Abou-Chakra, J., Kim, J., Drake, J., Peters, J., Schneider, J., Hsu, J., Bohg, J., Bingham, J., Wu, J., Gao, J., Hu, J., Wu, J., Wu, J., Sun, J., Luo, J., Gu, J., Tan, J., Oh, J., Wu, J., Lu, J., Yang, J., Malik, J., Silvério, J., Hejna, J., Booher, J., Tompson, J., Yang, J., Salvador, J., Lim, J.~J., Han, J., Wang, K., Rao, K., Pertsch, K., Hausman, K., Go, K., Gopalakrishnan, K., Goldberg, K., Byrne, K., Oslund, K., Kawaharazuka, K., Black, K., Lin, K., Zhang, K., Ehsani, K., Lekkala, K., Ellis, K., Rana, K., Srinivasan, K., Fang, K., Singh, K.~P., Zeng, K.-H., Hatch, K., Hsu, K., Itti, L., Chen, L.~Y., Pinto, L., Fei-Fei, L., Tan, L., Fan, L.~J., Ott, L., Lee, L., Weihs, L., Chen, M., Lepert, M., Memmel, M., Tomizuka, M., Itkina, M., Castro, M.~G., Spero, M., Du, M., Ahn, M., Yip, M.~C., Zhang, M., Ding, M., Heo, M., Srirama, M.~K., Sharma, M., Kim, M.~J., Kanazawa, N., Hansen, N., Heess, N., Joshi, N.~J., Suenderhauf, N., Liu, N., Palo, N.~D., Shafiullah, N. M.~M.,
  Mees, O., Kroemer, O., Bastani, O., Sanketi, P.~R., Miller, P.~T., Yin, P., Wohlhart, P., Xu, P., Fagan, P.~D., Mitrano, P., Sermanet, P., Abbeel, P., Sundaresan, P., Chen, Q., Vuong, Q., Rafailov, R., Tian, R., Doshi, R., Mart{'i}n-Mart{'i}n, R., Baijal, R., Scalise, R., Hendrix, R., Lin, R., Qian, R., Zhang, R., Mendonca, R., Shah, R., Hoque, R., Julian, R., Bustamante, S., Kirmani, S., Levine, S., Lin, S., Moore, S., Bahl, S., Dass, S., Sonawani, S., Song, S., Xu, S., Haldar, S., Karamcheti, S., Adebola, S., Guist, S., Nasiriany, S., Schaal, S., Welker, S., Tian, S., Ramamoorthy, S., Dasari, S., Belkhale, S., Park, S., Nair, S., Mirchandani, S., Osa, T., Gupta, T., Harada, T., Matsushima, T., Xiao, T., Kollar, T., Yu, T., Ding, T., Davchev, T., Zhao, T.~Z., Armstrong, T., Darrell, T., Chung, T., Jain, V., Vanhoucke, V., Zhan, W., Zhou, W., Burgard, W., Chen, X., Chen, X., Wang, X., Zhu, X., Geng, X., Liu, X., Liangwei, X., Li, X., Pang, Y., Lu, Y., Ma, Y.~J., Kim, Y., Chebotar, Y., Zhou, Y., Zhu, Y., Wu,
  Y., Xu, Y., Wang, Y., Bisk, Y., Dou, Y., Cho, Y., Lee, Y., Cui, Y., Cao, Y., Wu, Y.-H., Tang, Y., Zhu, Y., Zhang, Y., Jiang, Y., Li, Y., Li, Y., Iwasawa, Y., Matsuo, Y., Ma, Z., Xu, Z., Cui, Z.~J., Zhang, Z., Fu, Z., and Lin, Z.
\newblock Open {X-E}mbodiment: Robotic learning datasets and {RT-X} models.
\newblock \url{https://arxiv.org/abs/2310.08864}, 2023.

\bibitem[Cui et~al.(2022)Cui, Wang, Shafiullah, and Pinto]{cui2022play}
Cui, Z.~J., Wang, Y., Shafiullah, N. M.~M., and Pinto, L.
\newblock From play to policy: Conditional behavior generation from uncurated robot data.
\newblock \emph{arXiv preprint arXiv:2210.10047}, 2022.

\bibitem[Damen et~al.(2020)Damen, Doughty, Farinella, Fidler, Furnari, Kazakos, Moltisanti, Munro, Perrett, Price, et~al.]{damen2020epic}
Damen, D., Doughty, H., Farinella, G.~M., Fidler, S., Furnari, A., Kazakos, E., Moltisanti, D., Munro, J., Perrett, T., Price, W., et~al.
\newblock The epic-kitchens dataset: Collection, challenges and baselines.
\newblock \emph{IEEE Transactions on Pattern Analysis and Machine Intelligence}, 43\penalty0 (11):\penalty0 4125--4141, 2020.

\bibitem[Darcet et~al.(2023)Darcet, Oquab, Mairal, and Bojanowski]{darcet2023vitneedreg}
Darcet, T., Oquab, M., Mairal, J., and Bojanowski, P.
\newblock Vision transformers need registers, 2023.

\bibitem[Dass et~al.(2023)Dass, Yapeter, Zhang, Zhang, Pertsch, Nikolaidis, and Lim]{dass2023jacoplay}
Dass, S., Yapeter, J., Zhang, J., Zhang, J., Pertsch, K., Nikolaidis, S., and Lim, J.~J.
\newblock {CLVR} jaco play dataset, 2023.
\newblock URL \url{https://github.com/clvrai/clvr_jaco_play_dataset}.

\bibitem[Dosovitskiy et~al.(2021)Dosovitskiy, Beyer, Kolesnikov, Weissenborn, Zhai, Unterthiner, Dehghani, Minderer, Heigold, Gelly, Uszkoreit, and Houlsby]{dosovitskiy2021ViT}
Dosovitskiy, A., Beyer, L., Kolesnikov, A., Weissenborn, D., Zhai, X., Unterthiner, T., Dehghani, M., Minderer, M., Heigold, G., Gelly, S., Uszkoreit, J., and Houlsby, N.
\newblock An image is worth 16x16 words: Transformers for image recognition at scale, 2021.
\newblock URL \url{https://arxiv.org/abs/2010.11929}.

\bibitem[Du et~al.(2023)Du, Yang, Dai, Dai, Nachum, Tenenbaum, Schuurmans, and Abbeel]{du2023unipi}
Du, Y., Yang, M., Dai, B., Dai, H., Nachum, O., Tenenbaum, J.~B., Schuurmans, D., and Abbeel, P.
\newblock Learning universal policies via text-guided video generation, 2023.
\newblock URL \url{https://arxiv.org/abs/2302.00111}.

\bibitem[Ebert et~al.(2021)Ebert, Yang, Schmeckpeper, Bucher, Georgakis, Daniilidis, Finn, and Levine]{ebert2021bridge}
Ebert, F., Yang, Y., Schmeckpeper, K., Bucher, B., Georgakis, G., Daniilidis, K., Finn, C., and Levine, S.
\newblock Bridge data: Boosting generalization of robotic skills with cross-domain datasets.
\newblock \emph{arXiv preprint arXiv:2109.13396}, 2021.

\bibitem[Esser et~al.(2024)Esser, Kulal, Blattmann, Entezari, M{\"u}ller, Saini, Levi, Lorenz, Sauer, Boesel, et~al.]{esser2024scaling}
Esser, P., Kulal, S., Blattmann, A., Entezari, R., M{\"u}ller, J., Saini, H., Levi, Y., Lorenz, D., Sauer, A., Boesel, F., et~al.
\newblock Scaling rectified flow transformers for high-resolution image synthesis.
\newblock In \emph{Forty-first International Conference on Machine Learning}, 2024.

\bibitem[Goyal et~al.(2017)Goyal, Ebrahimi~Kahou, Michalski, Materzynska, Westphal, Kim, Haenel, Fruend, Yianilos, Mueller-Freitag, et~al.]{goyal2017something}
Goyal, R., Ebrahimi~Kahou, S., Michalski, V., Materzynska, J., Westphal, S., Kim, H., Haenel, V., Fruend, I., Yianilos, P., Mueller-Freitag, M., et~al.
\newblock The" something something" video database for learning and evaluating visual common sense.
\newblock In \emph{Proceedings of the IEEE international conference on computer vision}, pp.\  5842--5850, 2017.

\bibitem[Grauman et~al.(2022)Grauman, Westbury, Byrne, Chavis, Furnari, Girdhar, Hamburger, Jiang, Liu, Liu, et~al.]{grauman2022ego4d}
Grauman, K., Westbury, A., Byrne, E., Chavis, Z., Furnari, A., Girdhar, R., Hamburger, J., Jiang, H., Liu, M., Liu, X., et~al.
\newblock Ego4d: Around the world in 3,000 hours of egocentric video.
\newblock In \emph{Proceedings of the IEEE/CVF Conference on Computer Vision and Pattern Recognition}, pp.\  18995--19012, 2022.

\bibitem[Gupta et~al.(2023)Gupta, Wu, Deng, and Fei-Fei]{gupta2023siamMAE}
Gupta, A., Wu, J., Deng, J., and Fei-Fei, L.
\newblock Siamese masked autoencoders, 2023.
\newblock URL \url{https://arxiv.org/abs/2305.14344}.

\bibitem[He et~al.(2021)He, Chen, Xie, Li, Dollár, and Girshick]{he2021MAE}
He, K., Chen, X., Xie, S., Li, Y., Dollár, P., and Girshick, R.
\newblock Masked autoencoders are scalable vision learners, 2021.
\newblock URL \url{https://arxiv.org/abs/2111.06377}.

\bibitem[Heo et~al.(2023)Heo, Lee, Lee, and Lim]{heo2023furniturebench}
Heo, M., Lee, Y., Lee, D., and Lim, J.~J.
\newblock Furniturebench: Reproducible real-world benchmark for long-horizon complex manipulation.
\newblock In \emph{Robotics: Science and Systems}, 2023.

\bibitem[Hu et~al.(2023)Hu, Russell, Yeo, Murez, Fedoseev, Kendall, Shotton, and Corrado]{hu2023gaia1generativeworldmodel}
Hu, A., Russell, L., Yeo, H., Murez, Z., Fedoseev, G., Kendall, A., Shotton, J., and Corrado, G.
\newblock Gaia-1: A generative world model for autonomous driving, 2023.
\newblock URL \url{https://arxiv.org/abs/2309.17080}.

\bibitem[Jang et~al.(2022)Jang, Irpan, Khansari, Kappler, Ebert, Lynch, Levine, and Finn]{jang2022bc}
Jang, E., Irpan, A., Khansari, M., Kappler, D., Ebert, F., Lynch, C., Levine, S., and Finn, C.
\newblock Bc-z: Zero-shot task generalization with robotic imitation learning.
\newblock In \emph{Conference on Robot Learning}, pp.\  991--1002. PMLR, 2022.

\bibitem[Jin et~al.(2024)Jin, Sun, Xu, Xu, Chen, Jiang, Huang, Song, Liu, Zhang, Song, Gai, and Mu]{jin2024videolavit}
Jin, Y., Sun, Z., Xu, K., Xu, K., Chen, L., Jiang, H., Huang, Q., Song, C., Liu, Y., Zhang, D., Song, Y., Gai, K., and Mu, Y.
\newblock Video-lavit: Unified video-language pre-training with decoupled visual-motional tokenization, 2024.
\newblock URL \url{https://arxiv.org/abs/2402.03161}.

\bibitem[Kalashnikov et~al.(2018)Kalashnikov, Irpan, Pastor, Ibarz, Herzog, Jang, Quillen, Holly, Kalakrishnan, Vanhoucke, et~al.]{kalashnikov2018scalable}
Kalashnikov, D., Irpan, A., Pastor, P., Ibarz, J., Herzog, A., Jang, E., Quillen, D., Holly, E., Kalakrishnan, M., Vanhoucke, V., et~al.
\newblock Qt-opt: Scalable deep reinforcement learning for vision-based robotic manipulation.
\newblock In \emph{CoRL}, pp.\  651--673, 2018.

\bibitem[Karamcheti et~al.(2023)Karamcheti, Nair, Chen, Kollar, Finn, Sadigh, and Liang]{karamcheti2023voltron}
Karamcheti, S., Nair, S., Chen, A.~S., Kollar, T., Finn, C., Sadigh, D., and Liang, P.
\newblock Language-driven representation learning for robotics.
\newblock In \emph{Robotics: Science and Systems (RSS)}, 2023.

\bibitem[Khazatsky et~al.(2024)Khazatsky, Pertsch, Nair, Balakrishna, Dasari, Karamcheti, Nasiriany, Srirama, Chen, Ellis, Fagan, Hejna, Itkina, Lepert, Ma, Miller, Wu, Belkhale, Dass, Ha, Jain, Lee, Lee, Memmel, Park, Radosavovic, Wang, Zhan, Black, Chi, Hatch, Lin, Lu, Mercat, Rehman, Sanketi, Sharma, Simpson, Vuong, Walke, Wulfe, Xiao, Yang, Yavary, Zhao, Agia, Baijal, Castro, Chen, Chen, Chung, Drake, Foster, Gao, Herrera, Heo, Hsu, Hu, Jackson, Le, Li, Lin, Lin, Ma, Maddukuri, Mirchandani, Morton, Nguyen, O'Neill, Scalise, Seale, Son, Tian, Tran, Wang, Wu, Xie, Yang, Yin, Zhang, Bastani, Berseth, Bohg, Goldberg, Gupta, Gupta, Jayaraman, Lim, Malik, Martín-Martín, Ramamoorthy, Sadigh, Song, Wu, Yip, Zhu, Kollar, Levine, and Finn]{khazatsky2024droid}
Khazatsky, A., Pertsch, K., Nair, S., Balakrishna, A., Dasari, S., Karamcheti, S., Nasiriany, S., Srirama, M.~K., Chen, L.~Y., Ellis, K., Fagan, P.~D., Hejna, J., Itkina, M., Lepert, M., Ma, Y.~J., Miller, P.~T., Wu, J., Belkhale, S., Dass, S., Ha, H., Jain, A., Lee, A., Lee, Y., Memmel, M., Park, S., Radosavovic, I., Wang, K., Zhan, A., Black, K., Chi, C., Hatch, K.~B., Lin, S., Lu, J., Mercat, J., Rehman, A., Sanketi, P.~R., Sharma, A., Simpson, C., Vuong, Q., Walke, H.~R., Wulfe, B., Xiao, T., Yang, J.~H., Yavary, A., Zhao, T.~Z., Agia, C., Baijal, R., Castro, M.~G., Chen, D., Chen, Q., Chung, T., Drake, J., Foster, E.~P., Gao, J., Herrera, D.~A., Heo, M., Hsu, K., Hu, J., Jackson, D., Le, C., Li, Y., Lin, K., Lin, R., Ma, Z., Maddukuri, A., Mirchandani, S., Morton, D., Nguyen, T., O'Neill, A., Scalise, R., Seale, D., Son, V., Tian, S., Tran, E., Wang, A.~E., Wu, Y., Xie, A., Yang, J., Yin, P., Zhang, Y., Bastani, O., Berseth, G., Bohg, J., Goldberg, K., Gupta, A., Gupta, A., Jayaraman, D., Lim, J.~J.,
  Malik, J., Martín-Martín, R., Ramamoorthy, S., Sadigh, D., Song, S., Wu, J., Yip, M.~C., Zhu, Y., Kollar, T., Levine, S., and Finn, C.
\newblock Droid: A large-scale in-the-wild robot manipulation dataset.
\newblock \emph{arXiv preprint arXiv: 2403.12945}, 2024.

\bibitem[Kim et~al.(2024)Kim, Pertsch, Karamcheti, Xiao, Balakrishna, Nair, Rafailov, Foster, Lam, Sanketi, et~al.]{kim2024openvla}
Kim, M.~J., Pertsch, K., Karamcheti, S., Xiao, T., Balakrishna, A., Nair, S., Rafailov, R., Foster, E., Lam, G., Sanketi, P., et~al.
\newblock Openvla: An open-source vision-language-action model.
\newblock \emph{arXiv preprint arXiv:2406.09246}, 2024.

\bibitem[Kirillov et~al.(2023)Kirillov, Mintun, Ravi, Mao, Rolland, Gustafson, Xiao, Whitehead, Berg, Lo, Dollár, and Girshick]{kirillov2023segment}
Kirillov, A., Mintun, E., Ravi, N., Mao, H., Rolland, C., Gustafson, L., Xiao, T., Whitehead, S., Berg, A.~C., Lo, W.-Y., Dollár, P., and Girshick, R.
\newblock Segment anything, 2023.
\newblock URL \url{https://arxiv.org/abs/2304.02643}.

\bibitem[Ko et~al.(2023)Ko, Mao, Du, Sun, and Tenenbaum]{ko2023learningactactionlessvideos}
Ko, P.-C., Mao, J., Du, Y., Sun, S.-H., and Tenenbaum, J.~B.
\newblock Learning to act from actionless videos through dense correspondences, 2023.
\newblock URL \url{https://arxiv.org/abs/2310.08576}.

\bibitem[Li et~al.(2024)Li, Hsu, Gu, Pertsch, Mees, Walke, Fu, Lunawat, Sieh, Kirmani, Levine, Wu, Finn, Su, Vuong, and Xiao]{li24simpler}
Li, X., Hsu, K., Gu, J., Pertsch, K., Mees, O., Walke, H.~R., Fu, C., Lunawat, I., Sieh, I., Kirmani, S., Levine, S., Wu, J., Finn, C., Su, H., Vuong, Q., and Xiao, T.
\newblock Evaluating real-world robot manipulation policies in simulation.
\newblock \emph{arXiv preprint arXiv:2405.05941}, 2024.

\bibitem[Li et~al.(2018)Li, Liu, and Rehg]{li2018eye}
Li, Y., Liu, M., and Rehg, J.~M.
\newblock In the eye of beholder: Joint learning of gaze and actions in first person video.
\newblock In \emph{Proceedings of the European conference on computer vision (ECCV)}, pp.\  619--635, 2018.

\bibitem[Lin et~al.(2024)Lin, Cui, Xie, Hua, and Sadigh]{lin2024flowretrieval}
Lin, L.-H., Cui, Y., Xie, A., Hua, T., and Sadigh, D.
\newblock Flowretrieval: Flow-guided data retrieval for few-shot imitation learning, 2024.
\newblock URL \url{https://arxiv.org/abs/2408.16944}.

\bibitem[Liu et~al.(2023{\natexlab{a}})Liu, Nasiriany, Zhang, Bao, and Zhu]{liu2022robot}
Liu, H., Nasiriany, S., Zhang, L., Bao, Z., and Zhu, Y.
\newblock Robot learning on the job: Human-in-the-loop autonomy and learning during deployment.
\newblock In \emph{Robotics: Science and Systems (RSS)}, 2023{\natexlab{a}}.

\bibitem[Liu et~al.(2023{\natexlab{b}})Liu, Gong, and Liu]{liu2022flow}
Liu, X., Gong, C., and Liu, Q.
\newblock Flow straight and fast: Learning to generate and transfer data with rectified flow.
\newblock In \emph{International Conference on Learning Representations}, 2023{\natexlab{b}}.

\bibitem[Luo et~al.(2023)Luo, Xu, Geng, Feng, Fang, Tan, Schaal, and Levine]{luo2023multistage}
Luo, J., Xu, C., Geng, X., Feng, G., Fang, K., Tan, L., Schaal, S., and Levine, S.
\newblock Multi-stage cable routing through hierarchical imitation learning.
\newblock \emph{arXiv preprint arXiv:2307.08927}, 2023.

\bibitem[Luo et~al.(2024)Luo, Xu, Liu, Tan, Lin, Wu, Abbeel, and Levine]{luo2024fmb}
Luo, J., Xu, C., Liu, F., Tan, L., Lin, Z., Wu, J., Abbeel, P., and Levine, S.
\newblock Fmb: a functional manipulation benchmark for generalizable robotic learning.
\newblock \emph{arXiv preprint arXiv:2401.08553}, 2024.

\bibitem[Lynch et~al.(2023)Lynch, Wahid, Tompson, Ding, Betker, Baruch, Armstrong, and Florence]{lynch2023interactive}
Lynch, C., Wahid, A., Tompson, J., Ding, T., Betker, J., Baruch, R., Armstrong, T., and Florence, P.
\newblock Interactive language: Talking to robots in real time.
\newblock \emph{IEEE Robotics and Automation Letters}, 2023.

\bibitem[Ma et~al.(2023)Ma, Sodhani, Jayaraman, Bastani, Kumar, and Zhang]{ma2023vip}
Ma, Y.~J., Sodhani, S., Jayaraman, D., Bastani, O., Kumar, V., and Zhang, A.
\newblock Vip: Towards universal visual reward and representation via value-implicit pre-training, 2023.
\newblock URL \url{https://arxiv.org/abs/2210.00030}.

\bibitem[Majumdar et~al.(2023)Majumdar, Yadav, Arnaud, Ma, Chen, Silwal, Jain, Berges, Abbeel, Malik, Batra, Lin, Maksymets, Rajeswaran, and Meier]{vc2023}
Majumdar, A., Yadav, K., Arnaud, S., Ma, Y.~J., Chen, C., Silwal, S., Jain, A., Berges, V.-P., Abbeel, P., Malik, J., Batra, D., Lin, Y., Maksymets, O., Rajeswaran, A., and Meier, F.
\newblock Where are we in the search for an artificial visual cortex for embodied intelligence?
\newblock 2023.

\bibitem[Mandlekar et~al.(2018)Mandlekar, Zhu, Garg, Booher, Spero, Tung, Gao, Emmons, Gupta, Orbay, et~al.]{mandlekar2018roboturk}
Mandlekar, A., Zhu, Y., Garg, A., Booher, J., Spero, M., Tung, A., Gao, J., Emmons, J., Gupta, A., Orbay, E., et~al.
\newblock Roboturk: A crowdsourcing platform for robotic skill learning through imitation.
\newblock In \emph{Conference on Robot Learning}, pp.\  879--893. PMLR, 2018.

\bibitem[Mees et~al.(2023)Mees, Borja-Diaz, and Burgard]{mees2023grounding}
Mees, O., Borja-Diaz, J., and Burgard, W.
\newblock Grounding language with visual affordances over unstructured data.
\newblock In \emph{Proceedings of the IEEE International Conference on Robotics and Automation (ICRA)}, London, UK, 2023.

\bibitem[Mendonca et~al.(2023)Mendonca, Bahl, and Pathak]{mendonca2023structured}
Mendonca, R., Bahl, S., and Pathak, D.
\newblock Structured world models from human videos.
\newblock \emph{CoRL}, 2023.

\bibitem[Nair et~al.(2022)Nair, Rajeswaran, Kumar, Finn, and Gupta]{nair2022r3m}
Nair, S., Rajeswaran, A., Kumar, V., Finn, C., and Gupta, A.
\newblock R3m: A universal visual representation for robot manipulation, 2022.
\newblock URL \url{https://arxiv.org/abs/2203.12601}.

\bibitem[Nasiriany et~al.(2022)Nasiriany, Gao, Mandlekar, and Zhu]{nasiriany2022sailor}
Nasiriany, S., Gao, T., Mandlekar, A., and Zhu, Y.
\newblock Learning and retrieval from prior data for skill-based imitation learning.
\newblock In \emph{Conference on Robot Learning (CoRL)}, 2022.

\bibitem[Oquab et~al.(2023)Oquab, Darcet, Moutakanni, Vo, Szafraniec, Khalidov, Fernandez, Haziza, Massa, El-Nouby, Howes, Huang, Xu, Sharma, Li, Galuba, Rabbat, Assran, Ballas, Synnaeve, Misra, Jegou, Mairal, Labatut, Joulin, and Bojanowski]{oquab2023dinov2}
Oquab, M., Darcet, T., Moutakanni, T., Vo, H.~V., Szafraniec, M., Khalidov, V., Fernandez, P., Haziza, D., Massa, F., El-Nouby, A., Howes, R., Huang, P.-Y., Xu, H., Sharma, V., Li, S.-W., Galuba, W., Rabbat, M., Assran, M., Ballas, N., Synnaeve, G., Misra, I., Jegou, H., Mairal, J., Labatut, P., Joulin, A., and Bojanowski, P.
\newblock Dinov2: Learning robust visual features without supervision, 2023.

\bibitem[Pramanick et~al.(2023)Pramanick, Song, Nag, Lin, Shah, Shou, Chellappa, and Zhang]{pramanick2023egovlpv2}
Pramanick, S., Song, Y., Nag, S., Lin, K.~Q., Shah, H., Shou, M.~Z., Chellappa, R., and Zhang, P.
\newblock Egovlpv2: Egocentric video-language pre-training with fusion in the backbone.
\newblock In \emph{Proceedings of the IEEE/CVF International Conference on Computer Vision}, pp.\  5285--5297, 2023.

\bibitem[Quere et~al.(2020)Quere, Hagengruber, Iskandar, Bustamante, Leidner, Stulp, and Vogel]{quere_shared_2020}
Quere, G., Hagengruber, A., Iskandar, M., Bustamante, S., Leidner, D., Stulp, F., and Vogel, J.
\newblock Shared {Control} {Templates} for {Assistive} {Robotics}.
\newblock In \emph{2020 {IEEE} {International} {Conference} on {Robotics} and {Automation} ({ICRA})}, pp.\ ~7, Paris, France, 2020.

\bibitem[Radford et~al.(2021)Radford, Kim, Hallacy, Ramesh, Goh, Agarwal, Sastry, Askell, Mishkin, Clark, Krueger, and Sutskever]{radford2021CLIP}
Radford, A., Kim, J.~W., Hallacy, C., Ramesh, A., Goh, G., Agarwal, S., Sastry, G., Askell, A., Mishkin, P., Clark, J., Krueger, G., and Sutskever, I.
\newblock Learning transferable visual models from natural language supervision, 2021.
\newblock URL \url{https://arxiv.org/abs/2103.00020}.

\bibitem[Radosavovic et~al.(2022)Radosavovic, Xiao, James, Abbeel, Malik, and Darrell]{Radosavovic2022}
Radosavovic, I., Xiao, T., James, S., Abbeel, P., Malik, J., and Darrell, T.
\newblock Real-world robot learning with masked visual pre-training.
\newblock \emph{CoRL}, 2022.

\bibitem[Reed et~al.(2022)Reed, Zolna, Parisotto, Colmenarejo, Novikov, Barth-Maron, Gimenez, Sulsky, Kay, Springenberg, Eccles, Bruce, Razavi, Edwards, Heess, Chen, Hadsell, Vinyals, Bordbar, and de~Freitas]{reed2022generalistagent}
Reed, S., Zolna, K., Parisotto, E., Colmenarejo, S.~G., Novikov, A., Barth-Maron, G., Gimenez, M., Sulsky, Y., Kay, J., Springenberg, J.~T., Eccles, T., Bruce, J., Razavi, A., Edwards, A., Heess, N., Chen, Y., Hadsell, R., Vinyals, O., Bordbar, M., and de~Freitas, N.
\newblock A generalist agent, 2022.
\newblock URL \url{https://arxiv.org/abs/2205.06175}.

\bibitem[Rosete-Beas et~al.(2022)Rosete-Beas, Mees, Kalweit, Boedecker, and Burgard]{rosetebeas2022latent}
Rosete-Beas, E., Mees, O., Kalweit, G., Boedecker, J., and Burgard, W.
\newblock Latent plans for task agnostic offline reinforcement learning.
\newblock In \emph{Proceedings of the 6th Conference on Robot Learning (CoRL)}, 2022.

\bibitem[Saxena et~al.(2023)Saxena, Sharma, and Kroemer]{saxena2023multiresolution}
Saxena, S., Sharma, M., and Kroemer, O.
\newblock Multi-resolution sensing for real-time control with vision-language models.
\newblock In \emph{7th Annual Conference on Robot Learning}, 2023.
\newblock URL \url{https://openreview.net/forum?id=WuBv9-IGDUA}.

\bibitem[Schmidt \& Jiang(2023)Schmidt and Jiang]{schmidt2023learning}
Schmidt, D. and Jiang, M.
\newblock Learning to act without actions.
\newblock \emph{arXiv preprint arXiv:2312.10812}, 2023.

\bibitem[Shafiullah et~al.(2023)Shafiullah, Rai, Etukuru, Liu, Misra, Chintala, and Pinto]{shafiullah2023dobbe}
Shafiullah, N. M.~M., Rai, A., Etukuru, H., Liu, Y., Misra, I., Chintala, S., and Pinto, L.
\newblock On bringing robots home, 2023.

\bibitem[Shah et~al.(2023)Shah, Mart{\'\i}n-Mart{\'\i}n, and Zhu]{shah2023mutex}
Shah, R., Mart{\'\i}n-Mart{\'\i}n, R., and Zhu, Y.
\newblock {MUTEX}: Learning unified policies from multimodal task specifications.
\newblock In \emph{7th Annual Conference on Robot Learning}, 2023.
\newblock URL \url{https://openreview.net/forum?id=PwqiqaaEzJ}.

\bibitem[Team et~al.(2024)Team, Ghosh, Walke, Pertsch, Black, Mees, Dasari, Hejna, Kreiman, Xu, et~al.]{team2024octo}
Team, O.~M., Ghosh, D., Walke, H., Pertsch, K., Black, K., Mees, O., Dasari, S., Hejna, J., Kreiman, T., Xu, C., et~al.
\newblock Octo: An open-source generalist robot policy.
\newblock \emph{arXiv preprint arXiv:2405.12213}, 2024.

\bibitem[Walke et~al.(2023)Walke, Black, Lee, Kim, Du, Zheng, Zhao, Hansen-Estruch, Vuong, He, Myers, Fang, Finn, and Levine]{walke2023bridgedata}
Walke, H., Black, K., Lee, A., Kim, M.~J., Du, M., Zheng, C., Zhao, T., Hansen-Estruch, P., Vuong, Q., He, A., Myers, V., Fang, K., Finn, C., and Levine, S.
\newblock Bridgedata v2: A dataset for robot learning at scale, 2023.

\bibitem[Wang et~al.(2024)Wang, Zhu, Huang, Wang, Chen, and Lu]{wang2024worlddreamer}
Wang, X., Zhu, Z., Huang, G., Wang, B., Chen, X., and Lu, J.
\newblock Worlddreamer: Towards general world models for video generation via predicting masked tokens, 2024.
\newblock URL \url{https://arxiv.org/abs/2401.09985}.

\bibitem[Wu et~al.(2024)Wu, Yin, Feng, He, Li, Hao, and Long]{wu2024ivideogpt}
Wu, J., Yin, S., Feng, N., He, X., Li, D., Hao, J., and Long, M.
\newblock ivideogpt: Interactive videogpts are scalable world models.
\newblock \emph{arXiv preprint arXiv:2405.15223}, 2024.

\bibitem[Xiang et~al.(2024)Xiang, Liu, Gu, Gao, Ning, Zha, Feng, Tao, Hao, Shi, Liu, Xing, and Hu]{xiang2024pandorageneralworldmodel}
Xiang, J., Liu, G., Gu, Y., Gao, Q., Ning, Y., Zha, Y., Feng, Z., Tao, T., Hao, S., Shi, Y., Liu, Z., Xing, E.~P., and Hu, Z.
\newblock Pandora: Towards general world model with natural language actions and video states, 2024.
\newblock URL \url{https://arxiv.org/abs/2406.09455}.

\bibitem[Xiao et~al.(2022)Xiao, Radosavovic, Darrell, and Malik]{Xiao2022}
Xiao, T., Radosavovic, I., Darrell, T., and Malik, J.
\newblock Masked visual pre-training for motor control.
\newblock \emph{arXiv:2203.06173}, 2022.

\bibitem[Xu et~al.(2021)Xu, Dai, Liu, Gao, Lin, Qi, and Xiong]{xu2021STtransformer}
Xu, M., Dai, W., Liu, C., Gao, X., Lin, W., Qi, G.-J., and Xiong, H.
\newblock Spatial-temporal transformer networks for traffic flow forecasting, 2021.
\newblock URL \url{https://arxiv.org/abs/2001.02908}.

\bibitem[Yan et~al.(2023)Yan, Wu, and Wang]{ucsd_kitchens}
Yan, G., Wu, K., and Wang, X.
\newblock {ucsd kitchens Dataset}.
\newblock August 2023.

\bibitem[Yang et~al.(2024{\natexlab{a}})Yang, Du, Ghasemipour, Tompson, Kaelbling, Schuurmans, and Abbeel]{yang2024UniSim}
Yang, M., Du, Y., Ghasemipour, K., Tompson, J., Kaelbling, L., Schuurmans, D., and Abbeel, P.
\newblock Learning interactive real-world simulators, 2024{\natexlab{a}}.
\newblock URL \url{https://arxiv.org/abs/2310.06114}.

\bibitem[Yang et~al.(2024{\natexlab{b}})Yang, Walker, Parker-Holder, Du, Bruce, Barreto, Abbeel, and Schuurmans]{yang2024video}
Yang, S., Walker, J., Parker-Holder, J., Du, Y., Bruce, J., Barreto, A., Abbeel, P., and Schuurmans, D.
\newblock Video as the new language for real-world decision making, 2024{\natexlab{b}}.

\bibitem[Zheng et~al.(2024)Zheng, Peng, Yang, Shen, Li, Liu, Zhou, Li, and You]{opensora}
Zheng, Z., Peng, X., Yang, T., Shen, C., Li, S., Liu, H., Zhou, Y., Li, T., and You, Y.
\newblock Open-sora: Democratizing efficient video production for all, March 2024.
\newblock URL \url{https://github.com/hpcaitech/Open-Sora}.

\bibitem[Zhou et~al.(2023)Zhou, Dean, Srirama, Rajeswaran, Pari, Hatch, Jain, Yu, Abbeel, Pinto, Finn, and Gupta]{zhou2023train}
Zhou, G., Dean, V., Srirama, M.~K., Rajeswaran, A., Pari, J., Hatch, K., Jain, A., Yu, T., Abbeel, P., Pinto, L., Finn, C., and Gupta, A.
\newblock Train offline, test online: A real robot learning benchmark, 2023.

\bibitem[Zhu et~al.(2023{\natexlab{a}})Zhu, Tian, Xu, Ding, Zhan, and Tomizuka]{fanuc_manipulation2023}
Zhu, X., Tian, R., Xu, C., Ding, M., Zhan, W., and Tomizuka, M.
\newblock Fanuc manipulation: A dataset for learning-based manipulation with fanuc mate 200id robot.
\newblock 2023{\natexlab{a}}.

\bibitem[Zhu et~al.(2022)Zhu, Stone, and Zhu]{zhu2022bottom}
Zhu, Y., Stone, P., and Zhu, Y.
\newblock Bottom-up skill discovery from unsegmented demonstrations for long-horizon robot manipulation.
\newblock \emph{IEEE Robotics and Automation Letters}, 7\penalty0 (2):\penalty0 4126--4133, 2022.

\bibitem[Zhu et~al.(2023{\natexlab{b}})Zhu, Joshi, Stone, and Zhu]{zhu2023viola}
Zhu, Y., Joshi, A., Stone, P., and Zhu, Y.
\newblock Viola: Imitation learning for vision-based manipulation with object proposal priors, 2023{\natexlab{b}}.

\end{thebibliography}
\bibliographystyle{main}

\newpage
\appendix

\section{Dataset}
\label{appendix:dataset}

We present the datasets used for pre-training in Table \ref{table:dataset}. In total, these datasets comprise approximately 0.8 million robot trajectories and 2.0 million filtered human activity video clips. The robot data ratios are from \citep{team2024octo}.

\begin{table}[htb!]
\centering
\begin{tabular}{cc}\toprule  
Robot Dataset               & Mix Ratio (\%) \\\midrule  
Kuka   \citep{kalashnikov2018scalable}                     & 7.72          \\
Bridge \citep{walke2023bridgedata,ebert2021bridge}                     & 8.08          \\
Taco Play \citep{rosetebeas2022latent,mees2023grounding}                  & 1.82           \\
Jaco Play  \citep{dass2023jacoplay}                 & 0.24           \\
Berkeley Cable Routing \citep{luo2023multistage}     & 0.12           \\
Roboturk \citep{mandlekar2018roboturk}                   & 1.40           \\
Viola  \citep{zhu2023viola}                     & 0.55           \\
Berkely Autolab UR5 \citep{BerkeleyUR5Website}         & 0.73           \\
Toto \citep{zhou2023train}                       & 1.21           \\
Language Table  \citep{lynch2023interactive}             & 2.67           \\
Stanford Hydra Dataset \citep{belkhale2023hydra}      & 2.67           \\
Austin Buds Dataset \citep{zhu2022bottom}         & 0.12           \\
NYU Franka Play Dataset \citep{cui2022play}     & 0.49           \\
Furniture Bench Dataset  \citep{heo2023furniturebench}   & 1.46           \\
UCSD Kitchen Dataset \citep{ucsd_kitchens}       & 0.06           \\
Austin Sailor Dataset \citep{nasiriany2022sailor}       & 1.34           \\
Austin Sirius Dataset  \citep{liu2022robot}     & 1.03           \\
DLR EDAN Shared Control \citep{quere_shared_2020}     & 0.06           \\
IAMLab CMU Pickup Insert \citep{saxena2023multiresolution}    & 0.55           \\
UTAustin Mutex \citep{shah2023mutex}              & 1.34           \\
Berkeley Fanuc Manipulation \citep{fanuc_manipulation2023} & 0.43           \\
CMU Stretch \citep{mendonca2023structured}                & 0.12           \\
BC-Z \citep{jang2022bc}                       & 4.56           \\
FMB Dataset \citep{luo2024fmb}                & 4.31           \\
DobbE \citep{shafiullah2023dobbe}                    & 0.85           \\
DROID \citep{khazatsky2024droid}                      & 6.07           \\
Ego4D \citep{grauman2022ego4d}                       & 32.1           \\
Something-Something V2 \citep{goyal2017something}      & 9.5            \\
EPIC-KITCHENS  \citep{damen2020epic}             & 8.0            \\
EGTEA Gaze+   \citep{li2018eye}              & 0.4            \\ \bottomrule  
\end{tabular}\caption{ Dataset, mixture weights, and number of training examples after filtering in the pre-training stage in IGOR.}
\label{table:dataset}
\end{table}

\paragraph{Data Filtering}
We observed that video quality significantly affects the action model, particularly for human activities video. Excessive shakiness in videos can introduce visual changes between consecutive frames that are unrelated to the agent's actions.

We calculate the camera motion over the videos, and filter approximately 40\% percent of open-world video data. For the remaining data, we further stabilize the videos. Although we retain only 60\% percent of open-world video data, we find that the action model improves dramatically.

\paragraph{Frame Interval}
A noticeable amount of visual changes is crucial for our latent action model. If we select two frames that are too close in time, the agent may barely move, resulting in visual changes that are not significant enough for inferring meaningful actions. Conversely, if the frames are too far apart, the changes might be too large to model accurately. 
To address this issue, we tune the sampling interval. For robot data, we choose frames that are three intervals apart, using $s_t$ and $s_{t+3}$ as the image-goal pair. For real world videos, we control the sampling. For real world data, we control the sample interval to be within $[0.1s, 0.5s]$.
For the action and policy model, the context frames follow the same interval, ensuring that each pair of consecutive frames maintains this consistent spacing.

\newpage
\section{Training Details}
\vspace{-1mm}
\label{appendix:training_details}

\subsection{Latent Action Model Training}
\vspace{-1mm}

The latent action model uses an ST-transformer equipped with a frozen DINO-v2 pretrained ViT image encoder. 
The latent action model uses a patch size of 14, and a codebook with $N=4$ tokens and size $|C|=32$, each with an embedding size of $D=128$. We train the latent action model with batch size $B=512$, training iterations of $140$K steps, and learning rate of $1.5e-4$ with Adam optimizer.\vspace{-0.7mm}

\subsection{Foundation World Model Training}
\vspace{-1mm}

We start on the top of the OpenSora \citep{opensora} model with newly initialized projection layers. The foundation world model with batch size $B=12$, training iterations of $48$K, and learning rate of $1e-4$ with Adam optimizer. 
\vspace{-1mm}

\subsection{Foundation Policy Model and Low-Level Policy Model}
\vspace{-1mm}

The latent action model uses an ST-transformer equipped with a frozen DINO-v2 pretrained ViT image encoder, following the latent action model's image encoder. 
The foundation policy model consists of $12$ layers of spatial and temporal attentions, each with $12$ attention heads and hidden dimension as $768$ and a patch size of 14.
We use frozen CLIP features for text instructions. We pretrain the foundation policy model with batch size $B=128$, training iterations of $124$K, and learning rate of $1e-4$ with Adam optimizer. 

For the low-level policy model, we add extra layers on top of the foundation policy model. We use a sub-task length of $\tau=3$ for finetuning the low-level policy model on RT-1 dataset.
We finetune the low-level policy model with batch size $B=128$, training iterations of $32$K, and learning rate of $1e-4$ with Adam optimizer. 

\vspace{-0.7mm}

\section{Additional Ablation Results}
\vspace{-1mm}
\label{appendix:ablation}

\subsection{Dataset ablation for Latent Action Model}
\vspace{-1mm}
We compare two different settings for the pre-training dataset: only use the robotic dataset (robot data), and use both robotic and human activity dataset (mixed data). We evaluate the validation loss on the latent action model on RT-1 dataset, which is held out from the pretraining dataset and serves for OOD evaluation. Validation loss of the latent action model assesses the extent to which the IDM and FDM can jointly generate latent actions and recover goal states from these latent actions conditioned on states on the unseen dataset. The results are shown in Table~\ref{tab:abl_dataset}. 
We find that the OOD validation loss is greatly reduced by adding human activity dataset. This may be due to the diversity of human videos, which comprise real daily life environments with lots of diverse backgrounds and objects. These results demonstrate that it is promising to leverage human data for improving robot tasks under the IGOR framework. 

\vspace{-1mm}
\begin{table}[h!]
\centering
\begin{tabular}{cc}\\\toprule  
& Validation Loss \\\midrule
Robot Data & $0.145$ \\  \midrule
Mixed Data & $0.112$ \\  \bottomrule
\end{tabular}
\caption{Validation loss on held-out dataset (RT-1) with different training data.}\label{tab:abl_dataset}
\end{table} 
\vspace{-2mm}

\section{Additional Highlight of Contributions}
\vspace{-1mm}

We would like to especially acknowledge Pushi Zhang's contributions, including his involvement from the very beginning in shaping the initial ideas for the IGOR project, his valuable  support in experiment analysis and debugging, his dedicated efforts during the intense final stages of paper writing and refinement, and his crucial work in building the project webpage. We also would like to thank Kaixin Wang's valuable suggestions on paper title and great effort for preparing the well-designed paper template.


\end{document}